\PassOptionsToPackage{round}{natbib}
\documentclass{article}
\usepackage[preprint]{neurips_2024}

\usepackage{fontspec}
\newfontfamily\FBsym[Path=./]{DejaVuSans-sym-sub.ttf}
\newfontfamily\FBkr[Path=./]{NotoSansCJKkr-sub.otf}
\newcommand{\cmark}{{\FBsym ✓}}
\newcommand{\xmark}{{\FBsym ✗}}

\usepackage{booktabs}
\usepackage{tabularx}
\newcolumntype{Y}{>{\raggedright\arraybackslash}X}
\usepackage{longtable}
\usepackage{xltabular}
\usepackage{array}
\usepackage{graphicx}
\usepackage{amsmath}
\usepackage{amssymb}
\usepackage{xfrac}
\usepackage{enumitem}

\usepackage{calc}

\usepackage{placeins}
\usepackage{flafter}  
\usepackage{float}    
\usepackage{caption}
\makeatletter
\renewcommand{\@noticestring}{Preprint. Licensed under CC BY 4.0}
\makeatother

\usepackage[colorlinks=true,allcolors=blue,breaklinks=true]{hyperref}
\hypersetup{
  pdftitle={The Geometry of Emotion Representations Across Scale, Depth, and Instruction Tuning},
  pdfauthor={Adam Hollowell},
  pdfsubject={Mechanistic interpretability; emotion representations; scale; instruction tuning},
  pdfkeywords={interpretability, emotion vectors, Gemma 2, scale, instruction tuning, logit lens}}
\usepackage{url}

\title{The Geometry of Emotion Representations\\Across Scale, Depth, and Instruction Tuning}
\author{%
  Adam Hollowell\\
  University of North Carolina at Chapel Hill\\
  \texttt{adam.hollowell@unc.edu}
}

\begin{document}
\maketitle

\begin{abstract}
\hyphenpenalty=10000 \exhyphenpenalty=10000 
We measure the emotion geometry of six LLMs spanning Google's Gemma~2 family. Using an extraction pipeline established by Sofroniew et al.~(2026) on Claude Sonnet 4.5 and replicated by Hollowell (2026) on Gemma~2 27B base, we compare emotion geometry across base and instruction-tuned variants of Gemma~2~2B, 9B, and 27B. First, all six model cells exhibit the core emotion geometry. A linear map fit between two cells places held-out emotions nearest their own vectors in the target model. Across all fifteen model pairs, fit in one direction each, this failed exactly once, and the median rank was 1 in every pair. Linear maps refit on cleaned embedding-table vectors recover every held-out emotion at rank~1 in all thirty ordered pairs. Second, scale spreads the geometry's variance across more directions without weakening its tie to valence. The leading component's share of variance falls from 31.6 percent (2B) to 29.8 (9B) to 26.7 (27B) in the base models and from 31.9 (2B) to 29.2 (9B) to 27.3 (27B) in the tuned ones, while its correlation with human valence ratings stays between .71 and .78 in all six models. Third, we project each emotion vector onto the output vocabulary and ask at what depth the vector's own emotion word enters the top 10,000 of the 256,000 tokens. In the 2B and 9B models the median emotion word is there from layer 0, while in the 27B models it begins near chance, stays outside the top 10,000 for half the network, and enters at layer 23 in one step. Fourth, tuning leaves every registered analysis-layer measure within its floor. At every size, valence correlation rises and arousal correlation falls by amounts within their floors, the latter against prediction. PC2's variance share also falls, with a pooled decrease significant after correction. Mid-late layer stability exceeds its floor and traces to base 27B's final layer. Code, data, and audit reports are released.

\end{abstract}

\section{Introduction}

\citet{sofroniew2026emotion} extracted vectors for 171 emotions from Claude Sonnet 4.5 \citep{anthropic2025sonnet45} and found them organized via a strong valence axis, a weaker and more variable arousal axis, soft but stable emotion families, and high family cohesion. \citet{hollowell2026replicating} replicated that emotion structure in an open-weights model, Gemma 2 27B base, inheriting the extraction pipeline unchanged. How the core emotion geometry holds across scale and tuning, however, remains an open question.

\begin{figure}[htbp]
\centering
\includegraphics{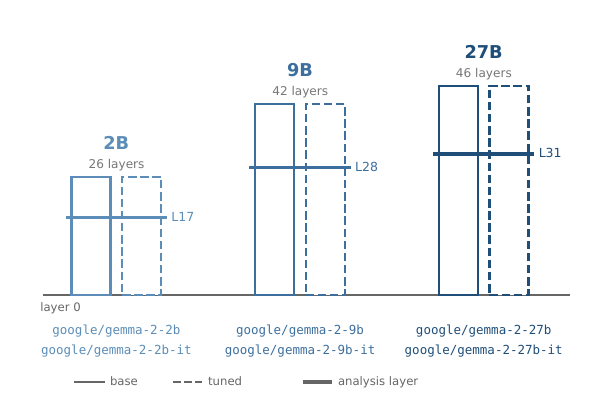}
\caption{\textbf{Model depths and analysis layers for base and instruction-tuned Gemma 2 at 2B, 9B, and 27B.}}
\label{fig:1}
\end{figure}

We apply the pipeline from \citet{sofroniew2026emotion} to the Gemma 2 family at three sizes, 2B, 9B, and 27B, each in base and instruction-tuned form, for six model cells. Figure~\ref*{fig:1} shows each cell's layer count, with its analysis layer marked at $\sfrac{2}{3}$ depth.

We define core emotion geometry as the structure common to both Claude Sonnet 4.5 in \citet{sofroniew2026emotion} and Gemma 2 27B base in \citet{hollowell2026replicating}, and we measure it in every cell as four properties: a valence axis, a weaker and more variable arousal axis, soft but stable family partitions, and family cohesion. We count a cell as showing the core emotion geometry when it passes a pre-stated test of the two axes, which requires that its valence and arousal correlations with human ratings are both significant by permutation (p $\leq$ .02 over 5,000 shuffles) and that the arousal correlation is the weaker of the two. We report family partitions and cohesion in every cell alongside base 27B's published values. The registration predicted directions for these measures but did not specify effect-size floors or pass criteria.

We find that all six Gemma 2 cells exhibit the core emotion geometry (\S4.1). The nine registered comparisons between cells find it closely shared, with pairwise-similarity agreement above .99 and family-partition agreement above its registered floor in all nine (\S4.2). A linear map fit between two cells on four fifths of the emotions places the median held-out emotion at retrieval rank 1 of 171 in the target cell (\S4.3).

Part of this geometry is present before any layer runs (\S5.1). Vectors built from each model's token-embedding table, with no forward pass and cleaned of generic token statistics, correlate moderately with human valence ratings in every cell (r = .48 to .51) and only weakly with arousal ratings (r = .12 to .18, significant in no cell). Linear maps refit on embedding-table vectors keep median rank 1 in every pair (\S5.1). Larger models place a smaller share of their emotion variance on the leading component, 31.6 percent at 2B against 26.7 at 27B in the base models, the opposite of our registered prediction, while its correlation with human valence ratings holds at .71 to .78 in all six cells (\S5.2).

In the 2B and 9B cells, each emotion vector's own word, projected onto the output vocabulary, already ranks at median in the top few thousand of 256,000 output tokens at layer 0, while in both 27B cells it starts near chance and stays outside the top 10,000 until layer 23, exactly half depth, where it enters the top 10,000 in a single step (\S6.2). The 27B geometry also consolidates at layer 23, though stricter readability thresholds are crossed later (\S6.3).

Instruction tuning leaves every registered analysis-layer measure within its floor (\S7.1). At every size, the valence correlation rises and the arousal correlation falls, the latter contrary to our prediction. PC2's variance share also falls at every size, and its pooled decrease remains significant after correction (Table~\ref*{tab:4}). The mid-late layer-stability contrast does exceed its floor, but \S8.3 traces that result to base 27B's final layer. Base and tuned activation profiles on neutral dialogues largely agree, though their finer differences are not preserved at 9B (\S7.2). Tuning's perturbation of the geometry grows across successive thirds of the network at every size, but the late rise at 27B depends on the final layer (\S7.3). Tuning also displaces some emotions much more than others (\S7.4).

We distinguish registered, pre-stated, and unregistered analyses according to when and how we specified them. Registered means fixed in the preregistration before any grid data existed, with the measure, the predicted direction, and the effect-size floor locked together. Pre-stated means the plan was fixed before it ran, by a hash or by a dated filing, but after the registered work, so it binds us without the preregistration's protection. Unregistered means none of these, and each unregistered analysis carries that label where its result is reported. Locked describes a plan fixed before running the relevant analyses, not necessarily at preregistration.

This study's code, results, and audit prompts and reports are released at \href{https://github.com/adamhollowell/emotion_probes/tree/main/gemma_grid}{\nolinkurl{github.com/adamhollowell/emotion_probes/tree/main/gemma_grid}}. The vectors, activations, per-story slices, and bootstrap files are hosted in the companion repository at \href{https://huggingface.co/adamhollowell/gemma-emotion-vectors}{\nolinkurl{huggingface.co/adamhollowell/gemma-emotion-vectors}}. \S11 lists what ships directly, what ships as recipes, and what can be re-derived from Gemma weights.

\section{Background and related work}

Earlier this year \citet{sofroniew2026emotion} presented the most extensive characterization of emotion representations in a large language model to date, extracting 171 emotion vectors from the residual stream of Claude Sonnet 4.5. \citet{hollowell2026replicating} replicated the representational core of the original study on base \texttt{google/gemma-2-27b} and showed that the model recovers the affective circumplex, the separation of valence and arousal, the family-level taxonomy, and a stable late-middle band containing their analysis layer at $\sfrac{2}{3}$ depth. The base-model recovery showed that the valence axis, in particular, predates alignment training and remains stable across emotion-set sizes and model depth. Appendix B maps the correspondence between this study and \citet{hollowell2026replicating}.

Several recent studies also examine emotion representations in Gemma models. \citet{choi2026latent} show representational similarity between Gemma-2-9B and Mistral-7B, though they probe classification-prompt activations rather than extract emotion vectors. \citet{keeman2026whether} uses supervised probes and activation patching, not extracted vectors, on eight emotions in base and tuned pairs of Llama-3.2-1B, Llama-3-8B, and Gemma-2-9B. \citet{presa2026compositional} study compositional emotions in Gemma 2 9B-IT and Llama 3.1 8B-Instruct, and \citet*{shu2026sae} study Gemma 3 at 4B and 12B, pretrained and tuned. Both read sparse-autoencoder features rather than difference-of-means vectors. \hyperlink{cite.vanderben2026}{Van der Ben et al.}~(\citeyear{vanderben2026}) use the method closest to ours. They test the generality of emotion structure in Gemma-4-E4B-it and Apertus-8B-Instruct-2509, using the 171 emotion concepts from \citet{sofroniew2026emotion} but with a smaller story corpus and neutral set.

\citet{soligo2026gemma} study Gemma from the behavioral side, evaluating expressions of distress in the outputs of Gemma 3 at 12B and 27B, base and instruction-tuned, and reducing them with preference optimization. \citet{zhang2025decoding} vary scale and fine-tuning across Qwen3 and Llama-3 and probe six emotions and a neutral class across a Reddit corpus of approximately 400,000 utterances. On scale, their probes separate emotions better in larger models, a result we take up in \S5.2. On fine-tuning, they probe a pre-trained model and a fine-tuned variant and find probe accuracy a few points apart, with emotion decodable from the pre-trained models at .69 to .76 (their Table 2). In relation to this study, they compare probe accuracy rather than geometry.

\citet*{tagliabue2026pain} extract a pain direction by difference of means from 25 models, including the six Gemma 2 checkpoints this study uses, and clean it by projecting out the principal components that explain half the variance of their control set. Scoring sentences on the pain direction separates painful sentences from matched controls with AUCs of .87 to 1.00 in every model across their two datasets, and the separation holds about equally from 2B to 72B and in base and tuned models. They measure one direction's separation, not the geometry of a set of emotion vectors.

\citet*{reichman2026emotions} ask the question closest to ours, though they do not test Gemma models. They extract emotion directions from synthetic sentences in base and tuned pairs of LLaMA-3.1-8B and OLMo-2, then ask whether human-written text from eight labeled datasets places the same emotions in the same relative positions. They report that tuned models reproduce the synthetic positions more closely than base models do, and that the distances between neighboring emotions are more distorted in the tuned models. Their one direct base-against-tuned comparison, rank correlations of emotion orderings along the top components, finds the orderings largely preserved under tuning (their Table 7). That finding is consistent in direction with our results in \S7.1, though on a smaller emotion set and without a significance test or a pre-set effect size.

No study to date measures emotion geometry across scale and tuning within the Gemma 2 family or across any family with the full pipeline of \citet{sofroniew2026emotion}. Appendix A sets out the design space and this study's relationship to its nearest neighbors, none of which combines a base arm, a readout over the model's whole output vocabulary, and a human-rating benchmark.

\section{Methods}

Following \citet{hollowell2026replicating}, we reproduce the extraction and characterization pipeline of \citet{sofroniew2026emotion}, resolving what it leaves unspecified by disclosed rules and changing only the subject models. Six Gemma 2 checkpoints (\hyperlink{cite.gemmateam2024gemma2}{Gemma Team}, \citeyear{gemmateam2024gemma2}) receive identical raw text: \texttt{google/gemma-2-2b}, \texttt{-2b-it}, \texttt{-9b}, \texttt{-9b-it}, \texttt{-27b}, and \texttt{-27b-it}, each pinned to a revision verified at weight-shard level and listed in Appendix I. Scale here names the three released sizes rather than parameter count alone, since the Gemma 2 report trains the 2B and 9B checkpoints by distillation from a larger model and the 27B checkpoint from scratch (\hyperlink{cite.gemmateam2024gemma2}{Gemma Team}, \citeyear{gemmateam2024gemma2}). \S9.2 states what that leaves unseparated.

\citet{sofroniew2026emotion} state their layer choice qualitatively (``a particular model layer about two-thirds of the way through the model'') and do not disclose Claude Sonnet 4.5's layer depth. Gemma 2 2B has 26 transformer layers, and our registered rule, round(2/3 of n\_layers), places the replication analysis layer at L17, since two-thirds of 26 is 17.33. Because 17.33 falls between two layers, we also registered L18 as a robustness check. The same rule gives L28 for 9B (42 layers) and L31 for 27B (46 layers).

We load each model through TransformerLens 3.3.0 \citep{nanda2022transformerlens} in the raw-weights regime, with no layer-norm folding and no weight centering, so the residual stream and unembedding correspond to the released weights (Appendix J). Extraction runs in bfloat16, following \citet{hollowell2026replicating}, and reads the post-residual stream at every layer. Gemma 2 applies its RMSNorm in the (1 + w) convention, which the logit-lens readout (\S6) handles explicitly rather than through weight folding. In what follows, we define layer 0 as the residual stream after the first transformer block. The embedding table is the static token-embedding lookup, read with no forward pass. Fractional depth is a layer's index divided by the model's layer count, so layer 23 of a 46-layer model sits at half depth.

Emotion vectors are extracted from a corpus of short stories, each written by Claude Sonnet 4.5 to depict one of 171 emotions from \citet{sofroniew2026emotion}. The original study did not release its story corpus, so we follow its topic-controlled design: 100 shared scenario topics, each rendered under all 171 emotions, 12 variants per emotion-topic pair, yielding 205,200 stories across 17,100 pairs. All stories were generated by \texttt{claude-sonnet-4-5-20250929} in one uniform regime, and no chat template is applied in any model cell. Confound cleanup uses 500 neutral dialogues generated under the same model and screened for emotional content. See \citet{hollowell2026replicating} for full pipeline details.

For each Gemma 2 model cell, we extract the emotion vectors following the three steps of \citet{sofroniew2026emotion}. We pool mean residual-stream activations from the 50th token onward at the analysis layer. We form difference-of-means vectors against the corpus grand mean. We then clean the vectors against the cell's own neutral-dialogue principal components to 50 percent cumulative variance. The set of components the cleaning removes is the cleanup basis, drawn for each cell from its own dialogues and never shared across cells. The same three steps, run on the embedding-table lookup for each token instead of the residual stream, build the embedding-table vectors of \S5.1 with no forward pass. We call what those components carry generic token statistics. Directions that dominate emotionally neutral text most likely encode frequency, length, and formatting, though we did not measure what they contain.

Base 27B was extracted again in this pipeline at its analysis layer, L31. Its rerun matched the locked vectors of \citet{hollowell2026replicating} in 24 of 24 checks, to about $1 \times 10^{-9}$. The six rows of Table~\ref*{tab:1} therefore come from one pipeline, and base 27B's row is a reproduction of a published measurement rather than a new model.

We estimate resampling uncertainty with four procedures and with the seed spread of the clustering. Elsewhere we use analytic Fisher standard errors, permutation tests, and a random-vector null, and we describe each where we use it. Story-level bootstraps draw 1,000 resamples per cell (Table~\ref*{tab:1}, \S4.2). A case bootstrap resamples the 171 emotions, at 2,000 draws, for the layer-stability contrasts of \S8.3. A second case bootstrap resamples the 45 rated emotions, at 2,000 draws with the same draw applied to every cell, for the dependence-aware tests of \S5.3 and the tuning intervals of \S7.1. The same procedure, at 1,000 draws per rating setting, resamples the rated emotions for the emotion-set control of \S8.2. Redraws of the cleanup basis repeat the neutral-dialogue draw and the cleaning, at 500 redraws for the readability analyses of \S6 and 1,000 for the displacement analysis of \S7.4.

Registered contrasts are Holm-corrected \citep{holm1979simple} within three registered families against floors locked before data. The correction holds each family's smallest p value to a threshold divided by the number of tests in the family, so that running several tests does not raise the chance of a false positive. We registered eleven rows for the tuning family (\S7.1), ten for the scale family (\S5, with the full battery in Appendix E), and one for the interaction family (\S7.1). Rows graded on their floor alone carry no p value and do not count toward the Holm correction. We also registered two vector-norm correlations but designated them exploratory, so they sit outside the families. The tuning battery of \S7.1 pools each base-to-tuned difference across the three sizes as the mean of the three per-size differences, a rule fixed at registration before any per-size value was visible, and the per-size differences always print beside the pooled one.

At registration we fixed the families, the signals, their directions, the floors, and the test for each kind of measure. Later, in the grading code, we fixed the rows in each family, the endpoint form of the scale contrast, the self-agreement test, and the resampling unit of the layer-stability bootstrap, after extraction and before grading any contrast, with each cell's own measured values in hand. Appendix C gives the dates and the concordance.

We locked every floor before any grid data existed. The .15 floor for correlation contrasts is one Fisher standard error at n = 45, treating the two correlations as independent ($\sqrt{2/42}$ = 0.22 in z, about .14 in r near .6), rounded up. The .10 floor for the two vector-norm correlations of \S7.1 is the same construction at n = 171. The .03 floor for neighbor cosine marks where the registration judged resampling wobble to end, in a bounded metric. The .01 floor for layer stability guards a quantity that reproduces to zero. The 1.0-point floor for variance shares comes from a one-point discrepancy between a caption and a figure in \citet{sofroniew2026emotion} \citep[\S5.1]{hollowell2026replicating}.

The Fisher tests treat the two correlations they compare as independent, although both are scored against the same emotions, so the .15 and .10 floors are stricter than a dependence-aware test would set. \S5.3 reports the two scale contrasts whose verdicts turn on a p value under both standard errors, and every tuning correlation contrast falls below its floor on size alone (\S7.1). The pooled tuning test makes a second independence assumption that errs in the opposite direction, treating the three size pairs as independent, and no tuning verdict rests on it (\S7.1, \S9.2).

Numbers follow the bounded-quantity convention: correlations, cosines, proportions, agreement indices, and p values cannot exceed 1 in absolute value and carry no leading zero, while unbounded quantities such as variance points and regression coefficients keep theirs. Differences are computed at full precision, so they will not always match the rounded values printed beside them.

\section{Core emotion geometry}

All six Gemma 2 cells exhibit the core emotion geometry (\S4.1). Three cross-cell checks find the geometry closely shared, with differences that stay small (\S4.2). A linear map fit between two cells predicts the vectors of emotions it never saw, placing the median held-out emotion nearest its own vector in the target cell (\S4.3).

\subsection{Emotion geometry across six Gemma 2 models}

In each cell we fit principal components on all 171 cleaned vectors, then correlate each emotion's position on the first two components with human ratings of valence and arousal for the 45 rated emotions from \citet{russell1977evidence}. The first component, PC1, captures 26.7 to 31.9 percent of the vectors' variance across the six cells (Table~\ref*{tab:1}), against about 27 percent in Claude Sonnet 4.5 \citep{sofroniew2026emotion}. PC1 tracks valence in every cell, with correlations of .711 to .777 against .81 for Claude Sonnet 4.5. The Claude Sonnet 4.5 values here and below are reference points, not comparisons. The registration tested no difference from them.

The second component, PC2, captures 12.7 to 13.9 percent of the variance, against about 14 percent in Claude Sonnet 4.5, and tracks arousal, with correlations of .363 to .668 against .66 for Claude Sonnet 4.5. Both axes are significant by permutation in every cell (p $\leq$ .02 over 5,000 shuffles). The arousal correlation is weaker than the valence correlation in every Gemma 2 cell, as \citet{hollowell2026replicating} observed for base 27B, and varies more between cells. Every cell therefore passes the test \S1 set, which asks that both axes be significant and that arousal be the weaker.

\begin{table}[htbp]
\footnotesize
\setlength{\tabcolsep}{4pt}
\caption{\textbf{Core emotion geometry in the six Gemma 2 cells.} The $\pm$ after each variance share and correlation is the standard deviation across 1,000 story-level bootstrap resamples, with each cell's cleanup basis held fixed. Self-agreement's $\pm$ is the standard deviation across 20 clustering seeds. Cohesion carries none, because it was registered with a direction but no test.}
\label{tab:1}
\begin{tabularx}{\linewidth}{@{}>{\hsize=0.933\hsize\raggedright\arraybackslash\hyphenpenalty=10000\exhyphenpenalty=10000}X>{\hsize=1.081\hsize\centering\arraybackslash}X>{\hsize=1.081\hsize\centering\arraybackslash}X>{\hsize=1.001\hsize\centering\arraybackslash}X>{\hsize=1.001\hsize\centering\arraybackslash}X>{\hsize=0.902\hsize\centering\arraybackslash}X>{\hsize=1.001\hsize\centering\arraybackslash}X@{}}
\toprule
\textbf{Cell} & \textbf{PC1 var (\%)} & \textbf{PC2 var (\%)} & \textbf{Valence r} & \textbf{Arousal r} & \textbf{Cohesion} & \textbf{Self-agreement} \\
\midrule
2B base & 31.62 $\pm$ 0.09 & 12.75 $\pm$ 0.06 & .711 $\pm$ .003 & .499 $\pm$ .021 & .706 & .717 $\pm$ .092 \\
2B tuned & 31.89 $\pm$ 0.09 & 12.66 $\pm$ 0.06 & .764 $\pm$ .003 & .363 $\pm$ .023 & .749 & .711 $\pm$ .075 \\
9B base & 29.85 $\pm$ 0.09 & 13.93 $\pm$ 0.06 & .715 $\pm$ .003 & .577 $\pm$ .009 & .748 & .695 $\pm$ .069 \\
9B tuned & 29.16 $\pm$ 0.08 & 13.57 $\pm$ 0.06 & .777 $\pm$ .003 & .554 $\pm$ .011 & .758 & .656 $\pm$ .083 \\
27B base & 26.73 $\pm$ 0.09 & 13.38 $\pm$ 0.06 & .720 $\pm$ .003 & .668 $\pm$ .010 & .756 & .695 $\pm$ .077 \\
27B tuned & 27.27 $\pm$ 0.08 & 12.85 $\pm$ 0.06 & .772 $\pm$ .003 & .578 $\pm$ .021 & .755 & .691 $\pm$ .078 \\
\bottomrule
\end{tabularx}
\end{table}

Following \citet{sofroniew2026emotion}, we also sort the 171 vectors into ten families with k-means and score each partition for cohesion and self-agreement. For each emotion we count how many of its ten nearest cosine neighbors fall inside its own family, and we average that fraction over the 171 emotions. Cohesion runs .706 to .758 across the six cells (Table~\ref*{tab:1}), so between seven and eight of an emotion's ten nearest neighbors share its family. Self-agreement is how closely partitions of the same cell from different random seeds agree, measured by the average adjusted Rand index between them (\S4.2), where 1 means the partitions are identical. Self-agreement runs .656 to .717 across the six cells (Table~\ref*{tab:1}). Together the two scores describe soft but stable families, in which neighbors mostly stay together while the boundaries between families shift from seed to seed. Cohesion and self-agreement fall in narrow ranges, and base 27B's values sit within both. The registration predicted their direction but set no floor or pass criterion for them, so they describe the cells rather than grade them.

\subsection{Cross-cell checks}

With emotion geometry established in each cell, we ask how closely the cells match one another at three levels: the pairwise similarities among the 171 emotion vectors, the plane of the two leading axes, and the partition of the emotions into ten families.

\subsubsection*{Pairwise similarities}
\addcontentsline{toc}{subsubsection}{Pairwise similarities}

Within each cell, the cosines between every pair of the 171 emotion vectors form a 171-by-171 table. We score how closely two cells' tables agree by the cosine between their entries, following representational similarity analysis \citep*{kriegeskorte2008rsa}, where 1 means identical. We make nine comparisons, the three tuning pairs (base and tuned at one size) and the six scale pairs (two sizes at one tuning level). A score below 1 could reflect a difference between the cells or only the sampling of their stories, so we also score how well each cell agrees with itself. For each emotion we resample its own 1,200 stories with replacement and recompute all 171 vectors from stored activations, which gives one replicate of the cell. We draw 1,000 replicates per cell, pair them into 500 pairs, and score each pair.

In every cell 95 percent of the replicate pairs agree at .9976 or higher, and we take .9976 as the reference for within-cell agreement under this resampling. Agreement between cells runs .9917 to .9964 across the nine comparisons, so every pair of cells differs by more than sampling noise, though every comparison still scores above .99. The six cells therefore carry nearly the same table of emotion similarities, and the small differences between them exceed what resampling the stories produces.

\subsubsection*{PC1 and PC2 planes}
\addcontentsline{toc}{subsubsection}{PC1 and PC2 planes}

Each cell's first two components, PC1 (valence) and PC2 (arousal), give each emotion a position in a plane. To compare two cells' planes, we rotate one cell's positions to fit the other's as closely as possible by orthogonal Procrustes alignment \citep{schonemann1966procrustes}, which \citet{ding2021grounding} find performs well for comparing models. The disparity is the mismatch left after the best rotation, and it is zero for planes identical up to rotation. Aligning the two replicates in each of the same 500 pairs gives the within-cell reference.

In every cell 95 percent of the replicate pairs align at a disparity of .0070 or less. Eight of the nine comparisons exceed .0070, with disparities of .016 to .049. The ninth, 9B tuned against 27B tuned, sits at .0048, within the replicate range. The best-fitting rotation never exceeds 2.2 degrees, and no alignment needs a reflection, which would mean one plane is a mirror image of the other. Across all nine comparisons, then, the two cells' planes line up almost without rotation, and the differences that remain lie in where individual emotions sit, not in how the axes are oriented. \S9.1 explains how a plane this stable can coexist with a weaker, more variable arousal axis.

\subsubsection*{Emotion families}
\addcontentsline{toc}{subsubsection}{Emotion families}

In each cell, k-means sorted the 171 vectors into ten families (\S4.1), and we compare two cells' partitions with the adjusted Rand index \citep{hubert1985comparing}. For every pair of emotions, it checks whether both partitions keep the two together or both separate them, and it corrects for the agreement chance would produce. A score of 1 means identical partitions, and a score near 0 means agreement no better than chance.

Agreement runs .53 to .81 across the nine comparisons. K-means starts from random points, so even one cell's partitions from different seeds agree only at .656 to .717 (Table~\ref*{tab:1}). Of the nine cross-cell scores, four fall below that range, two within it, and three above it. Families in different cells therefore match about as well as one cell's families match across seeds.

We registered .51 as the agreement below which we would call the families reorganized. The floor sits two standard deviations below the mean of a twenty-seed sweep on base 27B, run before any grid data existed, and it has not been revised. All nine comparisons clear it, and Appendix C explains why a floor recomputed from Table~\ref*{tab:1} would sit higher.

The three tests also checked a registered prediction, that tuning pairs would match more closely than scale pairs. At every level the three tuning pairs fall within the range of the six scale pairs, and the largest and smallest plane disparities both belong to scale pairs, so the prediction is not confirmed. Neither scale nor tuning consistently changes the shared geometry more than the other.

\subsection{Cross-cell prediction}

We now ask whether an emotion's vector in one cell can be recovered from the same emotion's vector in another. We use a Procrustes alignment between cells, which we call transport for short, and fit one linear map per pair of cells to predict each emotion's target vector from its source vector. The cells differ in width, 2,304 coordinates at 2B, 3,584 at 9B, and 4,608 at 27B, so their vectors cannot be compared directly. The map is deliberately rigid, a semi-orthogonal Procrustes map plus one global scale \citep{schonemann1966procrustes}. It applies one rotation, possibly with a reflection, and one rescaling to the whole set of vectors, with no stretch along any single direction and no shear. When the target is at least as wide as the source, the map preserves every angle among the vectors and scales every distance by one common factor. When the target is narrower, the map projects the vectors into the smaller space, which shrinks those distances.

We split the 171 emotions at random into five groups and fit each map five times. Each fit uses four groups, 136 or 137 emotions, and predicts the fifth, so every emotion is predicted once by a map fit without it. We score each prediction by the rank of the emotion's own vector among all 171 target vectors, including those the map was fit on. Rank 1 means the nearest target vector is the emotion's own, and the top-1 rate is the share of held-out emotions at rank 1. Held-out retrieval under a fitted map is the standard test in cross-lingual embedding alignment \citep{smith2017offline}. The map has no setting for any single emotion, so a prediction reaches rank 1 only when the two cells arrange the emotions largely alike. Because each map is fit only on emotion vectors, transport concerns the subspace they span, not the full representation space of either cell.

We designated one pair as the primary transport test before running, 2B base to 27B base, the grid's largest jump in scale and a doubling of width. The pass bar was fixed at the same time: for the primary pair, a median rank of at most 2 across all 171 predictions and a permutation p below .001. Every held-out emotion lands nearest its own vector from 2B base to 27B base, a top-1 rate of 1.0, and the median cosine between a mapped vector and its own target vector is .972. For the null we shuffled which target vector belongs to which emotion, refit, and scored top-1 accuracy, 10,000 times. No shuffle matched the observed score, which puts the add-one p at its smallest possible value, $9.999 \times 10^{-5}$.

\begin{figure}[htbp]
\centering
\includegraphics{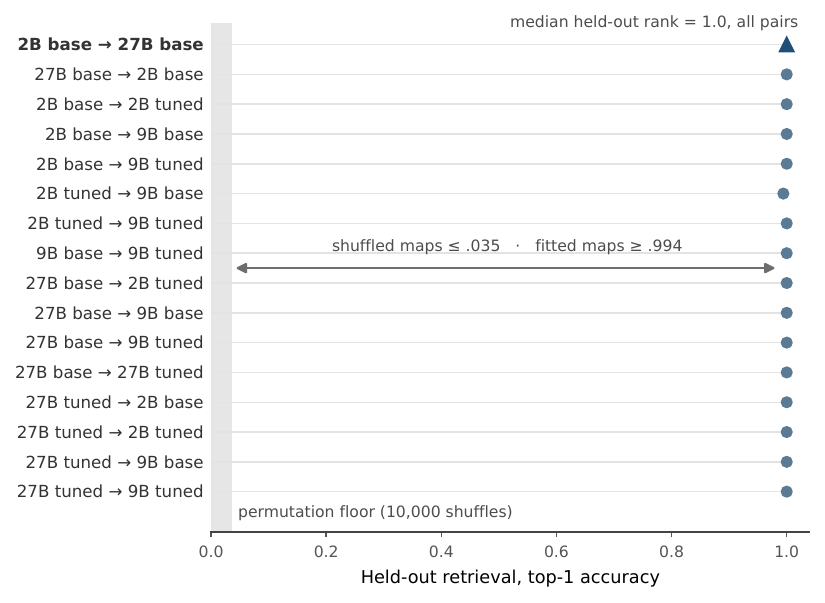}
\caption{\textbf{Held-out retrieval for all sixteen fits.} Each point is one fit, plotted at its top-1 rate, and every fit is plotted individually so no single pair can hide a failure. The band is the permutation null for 2B base to 27B base over 10,000 shuffled fits, and all sixteen fits sit far above it.}
\label{fig:2}
\end{figure}

We fit transport maps for all fifteen pairs of cells in one direction, and for 2B base to 27B base in the reverse direction as well, sixteen fits in all, and every fit has median rank 1 (Figure~\ref*{fig:2}). Across the fifteen pairs, one held-out emotion in one fit failed to land nearest its own vector, in the fit from 2B tuned to 9B base, a top-1 of 170 of 171.

Eight further checks were fixed before execution, and Appendix D reports each in full. Only one of them, truncating the source vectors below ten components, moved the median rank. Fit in the reverse direction, the maps add one miss, from 2B tuned to 27B base, again a top-1 of 170 of 171.

A pre-stated exploratory analysis asks whether transport depends on the two cells seeing the same stories. We split the 100 story topics into two fixed sets of 50 and built source vectors from one set and target vectors from the other, and all 32 cross-cell fits meet the median-rank criterion (Appendix D). A bag-of-words baseline with no fitted map retrieves as well on the same partition, at top-1 .877 and .883 in the two directions, unchanged when single-word emotion names are removed. Lexical information alone is therefore sufficient for strong performance on this retrieval task, which limits what transport can be read as showing.

\section{Embedding tables and scale effects}

This section asks how much of the core emotion geometry is already present in the token-embedding table, before any layer runs, and how scale changes the geometry's shape. Embedding-table vectors correlate moderately with human ratings of valence and weakly with arousal in every cell (\S5.1). Linear maps that recover held-out emotions at median rank 1 at $\sfrac{2}{3}$ depth (\S4.3), refit on embedding-table vectors, recover them at median rank 1 there too. Scale spreads the geometry's variance across more components without weakening the leading component's tie to valence (\S5.2). The arousal correlation rises from 2B to 27B, and so does the gap between the arousal correlation at the embedding table and at the analysis layer. Neither rise passes its registered test, and both depend on whether the 2B cells are read at L17, their analysis layer, or at L18, the registered alternative (\S5.3).

\subsection{Embedding tables track valence more than arousal}

\citet{hollowell2026replicating} introduced an embedding-table baseline that builds emotion vectors from the embedding table of base Gemma 2 27B and compares them with the vectors at the $\sfrac{2}{3}$-depth analysis layer. We repeat that baseline across all six cells to measure how much of the emotion geometry each model's embedding table carries before any layer runs.

For each story we average the embedding-table vectors over every token from the 50th onward (\S3). From those averages we build one difference-of-means vector per emotion and clean it against the neutral dialogues' own embeddings, removing their top three principal components, which carry half their variance in every cell. No forward pass runs at any point. We locked this analysis at registration, and for base 27B it reproduces the published baseline values exactly.

In all six cells, the embedding table correlates moderately with human valence ratings (r = .479 to .507, all p \textless{} .001) and weakly with arousal ratings (r = .125 to .179, all p $\geq$ .24, none significant). Vectors from each cell's analysis layer correlate more strongly with both (valence r = .711 to .777, arousal r = .363 to .668, Table~\ref*{tab:1}). From the embedding table to the analysis layer, the valence correlation rises by .21 to .29 within each cell and the arousal correlation by .24 to .49. The valence correlation is thus moderate before any layer runs, while the arousal correlation is weak.

Because cleaning removes generic token statistics (\S3), we also compare raw and cleaned vectors, as \citet{hollowell2026replicating} did for base 27B. On raw vectors the valence correlation is near zero at 2B and 9B (.029 to .074), which suggests that generic token statistics dominate the first component. In those cells the valence correlation emerges only on cleaned vectors. In the 27B cells the valence correlation is substantial on raw vectors as well (.378 and .380), possibly because the generic directions are proportionally weaker there or because the table puts proportionally more of its variance on valence. This study does not test which.

Cleaning does not simply inflate correlations, however. It lowers the arousal correlation in every cell, from .193 to .325 raw to .125 to .179 cleaned. Table~\ref*{tab:E.1} gives raw and cleaned values for every cell. Whether the analysis layer's valence axis is the table's own direction carried forward, or a different direction that tracks the same ratings, this study does not measure (\S10).

Emotion geometry beyond the valence correlation is present at the embedding table. In an unregistered check, we refit the linear maps of \S4.3 on the cleaned embedding-table vectors, for every pair of cells in both directions. Every fit kept median rank 1, and in every fit each held-out emotion landed nearest its own vector. Median cosines ran lower in every pair, .943 to .958 against .969 to .982 at the analysis layer, a gap of .02 to .03 with no uncertainty computed. The arrangement of emotions that one model's table shares with another's is present before the first layer runs.

\subsection{Scale spreads component variance}

To understand how scale changes the shape of the emotion geometry, we compare the share of variance on the first component across six cells. We registered a prediction that PC1's share of variance (Table~\ref*{tab:1}) would rise modestly from 2B to 27B, and its test, fixed at grading (\S3), compares the 2B and 27B endpoints against a floor of 1.0 point. This renders the 9B shares descriptive. (Table~\ref*{tab:E.4} records a second registered clause on this prediction, never graded.)

\begin{figure}[htbp]
\centering
\includegraphics{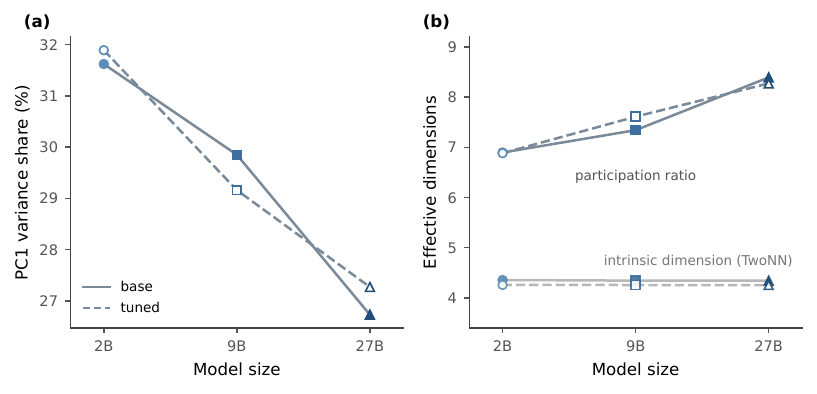}
\caption{\textbf{Component variance and dimension estimates across checkpoint sizes.} Panel (a) shows PC1's share of variance in each cell. Panel (b) shows two unregistered measures of effective dimension, the participation ratio and TwoNN. The TwoNN point estimates are similar across cells, but no uncertainty was computed for them.}
\label{fig:3}
\end{figure}

In the base chain, PC1's share falls from 31.62 percent at 2B to 29.85 at 9B to 26.73 at 27B (Figure~\ref*{fig:3}). In the tuned chain it falls from 31.89 to 29.16 to 27.27. The drop from base 2B to 27B is 4.88 percentage points, and the drop from tuned 2B to 27B is 4.62 percentage points, opposite to our prediction. Each chain's drop exceeds the floor more than fourfold and sits about 39 bootstrap standard errors from zero (38.6 and 39.4). Both are Holm-significant at p \textless{} $10^{-15}$, reported as a bound (Table~\ref*{tab:E.4}).

The drop is not an artifact of the cleanup. On raw vectors PC1's share still falls from 2B to 27B, by 4.8 points in the base chain and 5.5 in the tuned chain. On both raw and cleaned vectors, larger models put a smaller share of their emotion variance on the first principal component.

As PC1's share of variance falls, we track the participation ratio, which counts how many linear directions effectively share the variance. (If one component carried all the variance, the ratio would be 1, and if k components shared it equally, it would be k.) Here it rises with scale in both chains, from 6.90 to 7.34 to 8.39 in the base chain and from 6.88 to 7.61 to 8.27 in the tuned chain (Figure 3b). So as PC1's share falls, the variance is behaving as if spread over about one and a half additional components.

TwoNN \citep{facco2017estimating} estimates intrinsic dimension, the number of dimensions a point cloud locally spans, from each point's distances to its two nearest neighbors. For each cell's 171 vectors, TwoNN is about 4.3 (Figure 3b), so the vectors locally span about four dimensions at every size. Larger models therefore spread their emotion variance across more linear directions, while the local dimension of the vectors barely changes. Both are unregistered checks for which no uncertainty was computed.

A second registered contrast asks whether PC1's correlation with valence ratings changes from 2B to 27B, against a floor of .15. It does not. The correlation is .71 at 2B and .72 at 27B in the base chain, and .76 at 2B and .77 at 27B in the tuned chain (p = .928 and .927). A summary built on PC1 alone therefore describes less of a larger model's emotion variance, though PC1 itself is about as strongly correlated with valence at every size.

Finally, a smaller share of variance on PC1 leaves open whether larger models represent emotions less distinctly or only distribute them across more directions. \citet{zhang2025decoding} find that probes separate emotions better in larger Qwen3 and Llama-3 models, so we asked whether separability holds in this family. An unregistered nearest-class-mean classifier on held-out stories from our corpus picks the right emotion among the 171 about equally well at every scale (Appendix A). Larger models thus represent emotions no less distinctly, only across more directions. \S10 names an approach to identifying which distinctions among emotions those added directions carry.

\subsection{Scale may strengthen arousal}

We use two measures to follow arousal across scale. The first measure, arousal tracking, is the correlation of PC2 positions with human arousal ratings (Table~\ref*{tab:1}). Its registered test is a Fisher z comparison of the 2B and 27B endpoints within a chain, with a registered floor of .15. From 2B to 27B, arousal tracking rises by +.168 in the base chain and +.215 in the tuned chain, above the floor in both, and the test passes in neither (p = .237 and .200). The arousal axis gets better at ordering emotions by arousal as the model gets bigger, but not with enough certainty to count as a finding.

The second measure is the rise in the arousal correlation from the embedding table to the analysis layer, measured here in Fisher z units. We call this the depth gap. It has no registered floor, and its registered test is the same endpoint comparison as arousal tracking. From 2B to 27B the depth gap grows by +.203 in the base chain and +.229 in the tuned chain, and the test passes in neither (p = .511 and .459). As with arousal tracking, it moves in the registered direction without passing the registered test, so we report both results as leads rather than findings.

An unregistered bootstrap shows that the Fisher tests understate the evidence for both measures. The Fisher tests treat the correlations they compare as independent, although every correlation is scored against the same 45 rated emotions (\S3). The bootstrap resamples those 45 emotions once per round and recomputes every correlation on that same resample. The bootstrap's standard errors are 22 to 27 percent the size of the Fisher standard errors, so the Fisher tests are roughly four times too conservative here. Under the bootstrap, arousal tracking rises with p = $1.1 \times 10^{-5}$ in the base chain and $1.3 \times 10^{-6}$ in the tuned chain, and the depth gap grows with p = .003 and .001. Because the bootstrap was not registered, both measures remain not established and not shown absent. The bootstrap also resamples a selected set, the 45 emotions Russell and Mehrabian rated, rather than a random sample from a defined population. Its p values therefore show that both rises hold across those 45 emotions, not that they extend to emotions outside them. Story-level standard deviations in each cell's arousal correlation run only .009 to .023 (Table~\ref*{tab:1}), so the 45 rated emotions, not the story sample, set the limit on every test against ratings. Under the two broader rating sets of \S8.1, which cover most of the 171 emotions, the rise is smaller (Table~\ref*{tab:E.2}).

Because 2B's analysis layer is L17, rounded from 17.33, we ran analyses at L18 as a robustness check (\S3). Here that check is instructive, because on 2B base the arousal correlation is .499 at L17 and .656 at L18, a move of .157 that spans almost the whole of the +.168 rise the base chain shows from 2B to 27B. Reading 2B at L18, arousal tracking rises by only +.012 in the base chain and +.119 in the tuned chain, both below the .15 floor, and the same Fisher test is not significant in either chain (p = .92 and .45). Similarly, on 2B L18 the depth gap reverses sign in the base chain, from +.203 to $-$.034, and falls from +.229 to +.113 in the tuned chain. Under the bootstrap, the base chain's rise loses its support (p = .50) and the tuned chain's rise keeps its support (p = .0004), and the depth gap has p = .55 and .08. These checks do not re-grade the results from L17, but they flag that a scale contrast read at one layer per cell can turn on a one-layer rounding choice. For this reason, \S10 proposes scoring arousal over a depth band rather than at a single layer.

\section{Readability across scale}

\citet{hollowell2026replicating} found that emotion vectors in Gemma 2 27B base become readable in the output vocabulary at half depth. This section establishes the readout instrument (\S6.1) and asks when each cell's emotion words become readable (\S6.2). In the 2B and 9B cells the median emotion word reads from layer 0, while in the 27B cells it stays unreadable until layer 23, half depth. The 27B geometry consolidates at that same layer, and the two coincide where readability begins, at the top-10,000 line, but not under stricter readability lines (\S6.3).

\subsection{The readout instrument}

\citet[Table~1]{sofroniew2026emotion} use a vocabulary projection, the logit lens \citep{nostalgebraist2020logit}, to multiply an emotion vector by the model's unembedding matrix and read the highest-ranking tokens for twelve emotions: happy, inspired, loving, proud, calm, desperate, angry, guilty, sad, afraid, nervous, surprised. \citet[\S7.1 to \S7.2]{hollowell2026replicating} repeats the projection on Gemma 2 27B base and multiplies the cleaned vectors by the unembedding matrix alone, without the model's final normalization, since the normalized lens expects a residual state rather than a difference-of-means vector.

We extend this analysis to all six cells. We grade the vocabulary projection for the same twelve emotions under one rubric fixed before the data existed, which counts an emotion as on-theme when at least two of its top five tokens are emotion-related. A fresh-context Claude session read the rubric and each cell's top five tokens, with the rows in random order, and scored every row. We used its verdicts as returned, without regrading. To illustrate, Table~\ref*{tab:2} shows the guilty row in each cell.

\begin{table}[htbp]
\small
\caption{\textbf{Guilty's top five promoted tokens in each cell.} Asterisks mark tokens the grader judged emotion-related. Leading spaces are part of the token.}
\label{tab:2}
\begin{tabularx}{\linewidth}{@{}>{\hsize=0.479\hsize\raggedright\arraybackslash\hyphenpenalty=10000\exhyphenpenalty=10000}X>{\hsize=2.634\hsize\raggedright\arraybackslash}X>{\hsize=0.393\hsize\centering\arraybackslash}X>{\hsize=0.493\hsize\raggedright\arraybackslash}X@{}}
\toprule
\textbf{Cell} & \textbf{Top five tokens} & \textbf{Related} & \textbf{On-theme} \\
\midrule
2B base & \mbox{\texttt{' himself'}}, \mbox{\texttt{' nightmares'}}, \mbox{\texttt{' him'}}, \mbox{\texttt{' blames'}}*, \mbox{\texttt{' blamed'}}* & 2 & yes \\ \addlinespace[3pt]
2B tuned & \mbox{\texttt{' himself'}}, \mbox{\texttt{' him'}}, \mbox{\texttt{' nightmares'}}, \mbox{\texttt{' blamed'}}*, \mbox{\texttt{' damals'}} & 1 & no \\ \addlinespace[3pt]
9B base & \mbox{\texttt{'tagHelperRunner'}}, \mbox{\texttt{' deswegen'}}, \mbox{\texttt{' kvůli'}}, \mbox{\texttt{' desmotivaciones'}}, \mbox{\texttt{' deshalb'}} & 0 & no \\ \addlinespace[3pt]
9B tuned & \mbox{\texttt{' desmotivaciones'}}, \mbox{\texttt{'tagHelperRunner'}}, \mbox{\texttt{' hırka'}}, \mbox{\texttt{' zijne'}}, \mbox{\texttt{' saveiro'}} & 0 & no \\ \addlinespace[3pt]
27B base & \mbox{\texttt{' guilt'}}*, \mbox{\texttt{' carteles'}}, \mbox{\texttt{'irée'}}, \mbox{\texttt{'{\FBkr 졍}'}}, \mbox{\texttt{'etl'}} & 1 & no \\ \addlinespace[3pt]
27B tuned & \mbox{\texttt{' guilt'}}*, \mbox{\texttt{'irée'}}, \mbox{\texttt{'hamed'}}*, \mbox{\texttt{'{\FBsym ☍}'}}, \mbox{\texttt{' responsabilità'}}* & 3 & yes \\
\bottomrule
\end{tabularx}
\end{table}

Between seven and eleven of the twelve emotions grade on-theme in every cell (Table~\ref*{tab:3}). Four read on-theme in all six cells and nine in at least five, while guilty and calm set the floor, each for a different reason. As Table~\ref*{tab:2} shows for 27B base, guilty promotes the word guilt itself, but the other four of its top five are unrelated word fragments and tokens from other languages. Calm's top tokens are generically positive, which the rubric discounts for sharing an emotion's valence without naming it. These tallies are integer counts on a twelve-item rubric, and the section carries forward the family-wide claim, not any single cell's tally.

\begin{table}[htbp]
\small
\caption{\textbf{On-theme verdicts by emotion and cell.} \cmark{} marks an on-theme verdict and \xmark{} an off-theme one.}
\label{tab:3}
\begin{tabularx}{\linewidth}{@{}>{\hsize=2.113\hsize\raggedright\arraybackslash\hyphenpenalty=10000\exhyphenpenalty=10000}X>{\hsize=0.739\hsize\centering\arraybackslash}X>{\hsize=0.867\hsize\centering\arraybackslash}X>{\hsize=0.739\hsize\centering\arraybackslash}X>{\hsize=0.867\hsize\centering\arraybackslash}X>{\hsize=0.739\hsize\centering\arraybackslash}X>{\hsize=0.867\hsize\centering\arraybackslash}X>{\hsize=1.071\hsize\centering\arraybackslash}X@{}}
\toprule
\textbf{Emotion} & \textbf{2B base} & \textbf{2B tuned} & \textbf{9B base} & \textbf{9B tuned} & \textbf{27B base} & \textbf{27B tuned} & \textbf{Cells on-theme} \\
\midrule
happy & \cmark{} & \cmark{} & \cmark{} & \cmark{} & \cmark{} & \cmark{} & 6 \\
inspired & \cmark{} & \cmark{} & \xmark{} & \cmark{} & \cmark{} & \cmark{} & 5 \\
loving & \cmark{} & \cmark{} & \cmark{} & \cmark{} & \cmark{} & \xmark{} & 5 \\
proud & \cmark{} & \cmark{} & \cmark{} & \cmark{} & \cmark{} & \cmark{} & 6 \\
calm & \xmark{} & \xmark{} & \xmark{} & \xmark{} & \xmark{} & \cmark{} & 1 \\
desperate & \cmark{} & \cmark{} & \cmark{} & \cmark{} & \cmark{} & \cmark{} & 6 \\
angry & \cmark{} & \cmark{} & \cmark{} & \xmark{} & \cmark{} & \cmark{} & 5 \\
guilty & \cmark{} & \xmark{} & \xmark{} & \xmark{} & \xmark{} & \cmark{} & 2 \\
sad & \xmark{} & \xmark{} & \cmark{} & \cmark{} & \cmark{} & \cmark{} & 4 \\
afraid & \cmark{} & \xmark{} & \cmark{} & \cmark{} & \cmark{} & \cmark{} & 5 \\
nervous & \cmark{} & \xmark{} & \cmark{} & \cmark{} & \cmark{} & \cmark{} & 5 \\
surprised & \cmark{} & \cmark{} & \cmark{} & \cmark{} & \cmark{} & \cmark{} & 6 \\
Emotions on-theme & 10 & 7 & 9 & 9 & 10 & 11 & 56 of 72 \\
\bottomrule
\end{tabularx}
\end{table}

\subsection{When emotion words become readable}

We now ask when, in depth, emotion words become readable. The test here comes from \citet{hollowell2026replicating}, which defines readability onset as the layer where the median rank of each emotion's own word, among the 256,000 tokens its vector scores, crosses below 10,000, or roughly the top 4 percent of the vocabulary. A rank near the middle of the 256,000-token vocabulary means the word is promoted no better than chance. As a single-token test, it runs only on the 140 of the 171 emotions whose word is a single token (e.g., ``happy'') and not on those whose word has multiple tokens (e.g., ``grief-stricken''). No cross-cell direction was registered for this contrast, so the difference in scale below is descriptive.

The final normalization is a real step between the residual stream and the logits, so a test of what the output vocabulary can read should pass through it, unlike the projection of \S6.1. Following \citet[\S7.2 to \S7.3]{hollowell2026replicating}, we normalize in the (1 + w) form Gemma 2 uses (Appendix E). This is a vocabulary projection of an extracted direction, not of a residual state, so its ranks describe this projection and not the logits the model itself would produce at that layer.

\begin{figure}[htbp]
\centering
\includegraphics{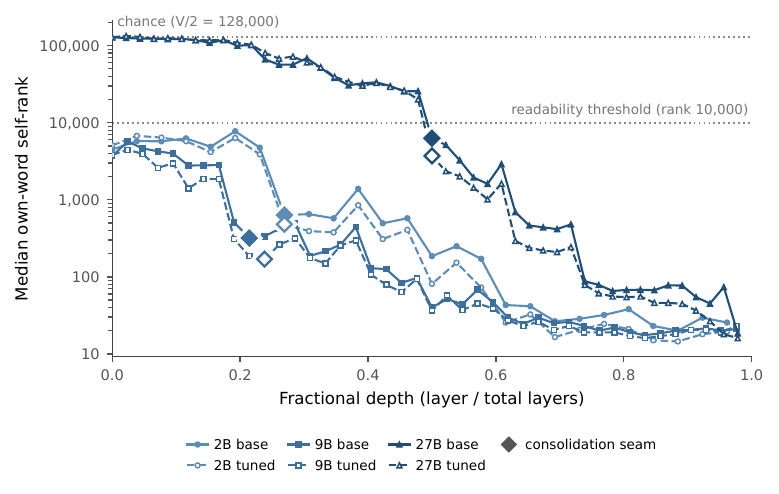}
\caption{\textbf{Median own-word rank by depth, with the consolidation seams marked.} Each curve is one cell's median rank for its own emotion word in the projected vocabulary, over the 140 single-token emotions. Seams (\S6.3) are defined at the .97 agreement criterion.}
\label{fig:4}
\end{figure}

In both 27B cells, the emotion words begin unreadable and stay unreadable for half the network (Figure~\ref*{fig:4}). The median rank starts nearly at chance, 127,700 in base and 128,600 in tuned, falls across the early and middle layers, and still sits near 25,500 in base and 26,000 in tuned at layer 21. At layer 23, exactly half depth, it drops in one step to about 6,300 in base and 3,700 in tuned, and the emotion words become readable. We call this run of layers across which emotion words are unreadable a gate, and it appears only in the 27B cells. The four smaller cells' median ranks sit between 3,700 and 5,100 at layer 0, already readable, and never rise above 7,750 at any layer. That is a claim about the median emotion, since per-emotion ranks at layer 0 spread widely in those cells and not every emotion reads early. Appendix F splits the same readout by valence and finds positive emotion words readable no later than negative ones in every cell, while negative words end the more readable in all six.

Both patterns, the late onset at 27B and the early readability at 2B and 9B, hold under resampling. Across 500 redraws of the cleanup basis, the onset stays at layer 23 in both 27B cells, and the four smaller cells read from layer 0 in every draw. We also passed random directions through the same readout, and they score near chance at all three widths, so the early readability comes from the emotion vectors themselves and not from the lower-dimensional residual streams of the smaller models. Appendix E reports these checks and the effect of removing the normalization and the cleanup in turn. Without the cleanup, 2B tuned no longer reads early.

The readability gate we observe here is the sharpest discontinuity by scale across the Gemma 2 family. At 2B and 9B the emotion words read from the first layer, yet at 27B they cannot be read for half the network and then become readable in a single step. We explore the gate further in \S6.3, but for now notice that this claim is about when the vocabulary projection can read an emotion word, not when the emotion's vector is present in the residual stream. Valence sits in the embedding table of both 27B cells, and transport maps refit on embedding-table vectors recover held-out emotions at median rank 1 (\S5.1, unregistered), yet the lens reads those cells' emotion words at chance at layer 0 and outside the top 10,000 until layer 23.

Further, logit-lens readouts degrade away from the final layer \citep{belrose2023tuned}, but that degradation does not explain the readability difference by scale. All six cells read through one operation, so a degradation that depended only on distance from the final layer would affect every cell alike at the same distance, and 9B, nearly as deep as 27B, reads from layer 0. Only 27B has a readability gate, so something specific to 27B's intermediate layers could be responsible, in the instrument or in the model. \S10 describes the test that would separate the two.

\subsection{The 27B gate at the consolidation seam}

If the 27B readout changes at layer 23, does the emotion geometry change there as well? To ask, we reuse the cosine between two 171-by-171 similarity tables (\S4.2), but within one cell rather than between cells. For each layer, we score how closely that layer's table agrees with the analysis layer's table (L31 for 27B). With this measure, \citet{hollowell2026replicating} observed a seam in 27B base from layers 22 to 26, where cross-layer similarity to L31 rose from .89 toward .999, with a seam-band mean of .978. For this unregistered measure we locate the consolidation seam at the first layer where that agreement reaches .97 and holds for three layers, so that it can be compared with the readability onset of \S6.2. The seam moves with the choice of .97, though across thresholds from .95 to .98 the 27B seam moves by at most one layer, and the smaller cells' seams stay between a seventh and about a third of depth (Appendix E).

In both 27B cells the consolidation seam sits at layer 23, exactly coincident with the readability onset at half of each cell's 46 layers. The smaller cells' seams come much earlier, at layer 7 in both 2B cells, 27 percent of depth, and at layer 9 in 9B base and layer 10 in 9B tuned, 21 and 24 percent. Under the 500 redraws of the cleanup basis from \S6.2, the 27B seam stays at layer 23 or 24. The distance between seam and onset is zero in 89.6 percent of draws in base and 75.6 percent in tuned, never exceeding one layer.

We pre-stated two tests to calibrate our analysis of readability and consolidation coinciding (Appendix E). The first replaces the onset line of \S6.2 with a rank taken from each cell's own median-rank trajectory, about 2,300 in 27B base and 1,700 in 27B tuned, four to six times stricter than 10,000. The rule fixed in advance counts a crossing that rounds to layers 22 to 25 as coincident with the seam. Both cells cross that line between layers 25 and 26, two to three layers past the seam. Rounded to the nearest layer, the tuned cell's crossing is 25 and counts as coincident, and the base cell's is 26 and counts as noncoincident, though the two differ by a fifth of a layer.

The second test tightens the readability line from the top 10,000 tokens to the top 1,000 and the top 100, and fixes in advance a window of layers 18 to 28 in which a crossing would count as coincident. Both cells first cross 1,000 at layer 29 and 100 at layer 34, six and eleven layers after the seam, so neither passes the test.

What survives, then, is a remarkable, but conditional, coincidence. In both 27B cells, and in none of the four smaller cells, the layer at which the geometry consolidates is the layer at which emotion words first become readable, and the two stay within a layer of each other across every redraw of the cleanup basis. Under stricter conditions, particularly as the readability measure tightens, the phenomena separate.

\section{Tuning effects}

This section asks what instruction tuning changes in the core emotion geometry. At $\sfrac{2}{3}$ depth, tuning leaves every registered measure within its floor (\S7.1). However, at every size the valence correlation rises and the arousal correlation falls, by less than their floors and contrary to our prediction for arousal. Base and tuned activation profiles on neutral dialogues largely agree, though their finer differences are not preserved at 9B (\S7.2). Tuning's perturbation of the geometry grows across successive thirds of the network at every size, but the late rise at 27B depends on the final layer (\S7.3). Tuning also displaces some emotions much more than others, and that ranking agrees across sizes (\S7.4).

\subsection{The tuning battery}

Here we test base-to-tuned differences in emotion geometry, beginning with valence and arousal correlations. We registered predictions that instruction tuning would raise both correlations.

Pooled across the three sizes, tuning raises the valence correlation by +.056 and changes the arousal correlation by $-$.083, against a registered floor of .15 for each (p = .32 and .36). The rise in valence correlation meets our predicted direction but is neither significant under the registered test nor above the floor, and the arousal change has the wrong sign for our prediction, at every size as well as pooled (Table~\ref*{tab:4}). Every other analysis-layer measure with a registered floor stays within it, at every size and pooled.

\begin{table}[htbp]
\small
\caption{\textbf{The tuning battery.} $\Delta$ is tuned minus base, and pp is percentage points. Pooled $\Delta$ is the mean of the three per-size $\Delta$ values. The p values are nominal. The two layer-stability rows carry no p and are printed at the precision of their artifacts (\S8.3).}
\label{tab:4}
\begin{tabularx}{\linewidth}{@{}>{\hsize=2.270\hsize\raggedright\arraybackslash\hyphenpenalty=10000\exhyphenpenalty=10000}X>{\hsize=0.772\hsize\centering\arraybackslash}X>{\hsize=0.772\hsize\centering\arraybackslash}X>{\hsize=0.772\hsize\centering\arraybackslash}X>{\hsize=0.772\hsize\centering\arraybackslash}X>{\hsize=0.581\hsize\centering\arraybackslash}X>{\hsize=1.062\hsize\centering\arraybackslash}X@{}}
\toprule
\textbf{Registered measure} & \textbf{$\Delta$ 2B} & \textbf{$\Delta$ 9B} & \textbf{$\Delta$ 27B} & \textbf{Pooled $\Delta$} & \textbf{Floor} & \textbf{p} \\
\midrule
Valence r (PC1 with valence ratings) & +.0535 & +.0620 & +.0520 & +.0558 & .15 & .320 \\
Arousal r (PC2 with arousal ratings) & $-$.1362 & $-$.0233 & $-$.0892 & $-$.0829 & .15 & .357 \\
PC1 variance share (pp) & +0.2706 & $-$0.6908 & +0.5362 & +0.0387 & 1.0 & .584 \\
PC2 variance share (pp) & $-$0.0865 & $-$0.3576 & $-$0.5258 & $-$0.3233 & 1.0 & $1.8 \times 10^{-10}$ \\
PC3 variance share (pp) & $-$0.2049 & +0.3829 & +0.2856 & +0.1545 & 1.0 & .0013 \\
Off-diagonal, PC1 with arousal ratings & +.0316 & +.0334 & +.0579 & +.0409 & .15 & .743 \\
Off-diagonal, PC2 with valence ratings & $-$.0807 & $-$.0364 & $-$.1471 & $-$.0881 & .15 & .472 \\
Neighbor cosine & +.0012 & +.0001 & +.0059 & +.0024 & .03 & .017 \\
Partition self-agreement & $-$.0058 & $-$.0399 & $-$.0041 & $-$.0166 & none & .798 \\
Mid-late layer stability & +.00023 & $-$.00001 & +.04236 & +.01419 & .01 & n/a \\
Layer stability, all layers & $-$.0063 & $-$.0062 & +.0146 & +.0007 & .01 & n/a \\
\bottomrule
\end{tabularx}
\end{table}

Two variance-share changes survive the Holm correction but remain below their registered floors. The second component's share falls 0.323 points pooled (p = $1.8 \times 10^{-10}$), at every size and by more at each larger size, and the third component's share rises 0.155 points (p = .0013), against a 1.0-point floor for each. Neighbor cosine, the cosine between nine emotion pairs fixed in advance from base 27B, also rises, by .0024 against a .03 floor, but its nominal p = .017 does not survive the correction. We grade all three null because each falls below its floor.

The two off-diagonal rows of Table~\ref*{tab:4} measure how much the valence axis picks up arousal ratings (PC1 with arousal) and how much the arousal axis picks up valence ratings (PC2 with valence), and the registered question is whether tuning changes these cross-correlations. Each axis is sign-flipped to agree with its own rating, so the valence and arousal correlations are non-negative by construction, while the off-diagonals, which reuse those signed axes, can be negative. Neither off-diagonal moves beyond its floor under tuning.

The leading component's variance share moves little, and partition self-agreement, which has no floor, moves $-$.017 (p = .80). The registration also listed three measures that Table~\ref*{tab:4} does not carry, because it excluded them from the Holm correction in advance. Two are norm correlations, which relate each emotion's vector length to its position on an axis and were registered as exploratory, and tuning leaves both under their .10 floor. The third is cohesion, registered with a direction but no floor, and it moves little (Appendix E).

One registered measure of the geometry across layers does not stay within its floor. Mid-late layer stability, the agreement of the emotion geometry among the sampled layers at or past half depth, rises under tuning by more than its floor, carried by the 27B pair. Its test is a case bootstrap fixed at grading (\S3), which we report in \S8.3 and trace to base 27B's final layer. The all-layer contrast stays within its floor pooled (+.0007), although its 27B value (+.0146) exceeds it.

We also registered one test of whether tuning's effect depends on scale, predicting that tuning would raise the arousal correlation more at 27B than at 2B. Arousal falls at both sizes (Table~\ref*{tab:4}), and the 27B delta exceeds the 2B delta by +.047 only because it falls less there. The difference is about a third of the .15 floor, and the Fisher z test of the contrast gives p = .94, so we find no support for the predicted dependence.

The .15 floor on the correlation rows inherits the independence assumption of the Fisher tests (\S3), which makes it six to fifteen times stricter than it would be if the standard errors allowed for the emotions both correlations share (Table~\ref*{tab:E.5}). An unregistered bootstrap that allows for that dependence puts all four pooled correlation shifts on one side of zero (Table~\ref*{tab:E.5}). With the two variance rows reliably nonzero as well, tuning moves the geometry by small, consistent amounts, and the verdicts here record differences below a floor and not differences shown to be absent.

These nulls are compatible with \citet{zhang2025decoding}, who find substantial emotion decodability before fine-tuning, though their measure is probe accuracy rather than emotion-vector geometry. \S9.1 relates the nulls to persona-selection accounts of post-training.

\subsection{Activation profiles are largely preserved, but 27B's shifts diverge}

The geometry measured in \S7.1 describes where the emotions sit relative to one another. It does not describe what each emotion's vector reads from the model's activations on neutral text, and tuning could change that without moving the geometry. \citet{sofroniew2026emotion} measured this for Claude by reading base and post-trained activations with one fixed set of probes, and we do the same at each size.

A probe here is an emotion's raw vector from \S3, before cleaning. Its reading on a dialogue is the cosine between the vector and the model's activation on that dialogue, so a high reading means the activation points the way the emotion's vector does. For each size we take the tuned cell's 171 probes and apply them to both the base and the tuned model on the same 500 neutral dialogues (\S3). An emotion's reading for a cell is the mean of its per-dialogue readings at the analysis layer, and a cell's 171 readings form its activation profile. The analysis is unregistered. It is an analog of the comparison in \citet{sofroniew2026emotion}, not a replication, because their prompts and token position differ from ours (Appendix G).

Base and tuned activation profiles agree almost completely, correlating across the 171 emotions at .9967 at 2B, .9993 at 9B, and .9987 at 27B. Much of that agreement comes from the contribution of each cell's mean activation to the probe readings.

Subtracting each cell's mean activation before the cosines are taken leaves what varies from dialogue to dialogue. On that centered reading the profiles correlate at .833 at 2B, .954 at 27B, and $-$.121 at 9B, where a few high-variance dimensions carry the disagreement (Appendix G). Base and tuned profiles therefore agree closely overall at every size. After centering, substantial agreement remains at 2B and 27B but not at 9B.

We define the activation shift as the tuned profile minus the base profile, using the original, uncentered readings, one value per emotion at each size. The shifts do not agree across sizes. The 2B and 9B shifts correlate at .54, but the 27B shift correlates with the 2B and 9B shifts at only .23 and .17, so the 27B shift profile agrees only weakly with those of the smaller models. The geometry's changes in \S7.1 show no such pattern by size.

At 27B the largest shifts in each direction gather around a few themes. The ten emotions tuning lowers most are nostalgic, regretful, remorseful, guilty, sympathetic, reflective, sentimental, self-critical, jealous, and sorry, and most of that group is guilt-related (regretful, remorseful, guilty, sorry, self-critical) or wistful (nostalgic, reflective, sentimental). The ten it raises most are resigned, docile, indifferent, nervous, patient, sensitive, anxious, uneasy, unnerved, and mad, and the group includes resigned, docile, and indifferent alongside four anxiety-related emotions, nervous, anxious, uneasy, and unnerved. \S7.4 finds three of the same emotions, regretful, remorseful, and guilty, among those tuning displaces furthest in the geometry, by a separate computation, and the study's one out-of-sample prediction was drawn from the lowered group (Appendix G). The Gemma shifts are at most weakly related to the per-emotion shifts \citet{sofroniew2026emotion} report for Claude Sonnet 4.5 (Appendix G).

\subsection{Tuning's perturbation grows with depth}

\citet[Figure~84]{sofroniew2026emotion} report from probe activations that post-training changes Claude's emotion representations more at mid-late layers than at early ones. We examine this question across the Gemma 2 family with unregistered exploratory analyses of the geometry and of the activation shifts defined in \S7.2. At each layer, a cell has a table of cosine similarities between every pair of distinct emotions among the 171. We correlate a base cell's table with its tuned cell's table at the same layer and define perturbation as one minus that Pearson correlation. Perturbation measures disagreement in the pattern of pairwise similarities. Pearson centers each table, so a shift in the overall level of the similarities, which the uncentered cosine of \S4.2 would register, does not enter. We divide each model's layers into three consecutive thirds of near-equal size, with any leftover layers added one each to the earliest thirds, and average within each third (Table~\ref*{tab:5}).

Perturbation rises from the early to the middle to the late third at every size. At 27B the late rise comes from the last two layers. Across the 27B pair, perturbation is .0056 at layer 43, then .0882 at layer 44 and .1755 at layer 45, the final layer, where \S8.3 documents base 27B as divergent by a separate measure. Without layers 44 and 45 the 27B late third falls below its early third (Table~\ref*{tab:5}, Appendix G). Geometric perturbation therefore rises across successive thirds at every Gemma 2 size, although the late rise at 27B depends on the final layer. This resembles Claude's reported depth pattern, measured there through probe activations.

\begin{table}[htbp]
\small
\caption{\textbf{Mean perturbation and mean activation shift, by third of each model's layers.} Layers in each third: 2B 0 to 8, 9 to 17, 18 to 25. 9B 0 to 13, 14 to 27, 28 to 41. 27B 0 to 15, 16 to 30, 31 to 45. The three further 27B rows drop the named layers from the late third.}
\label{tab:5}
\begin{tabularx}{\linewidth}{@{}>{\hsize=0.892\hsize\raggedright\arraybackslash\hyphenpenalty=10000\exhyphenpenalty=10000}X>{\hsize=2.858\hsize\raggedright\arraybackslash}X>{\hsize=0.394\hsize\centering\arraybackslash}X>{\hsize=0.480\hsize\centering\arraybackslash}X>{\hsize=0.376\hsize\centering\arraybackslash}X@{}}
\toprule
\textbf{Measure} & \textbf{Cell} & \textbf{Early third} & \textbf{Middle third} & \textbf{Late third} \\
\midrule
Perturbation & 2B & .0042 & .0055 & .0075 \\
Perturbation & 9B & .0034 & .0055 & .0073 \\
Perturbation & 27B & .0100 & .0142 & .0213 \\
Perturbation & 27B without layer 45 & .0100 & .0142 & .0103 \\
Perturbation & 27B without layer 44 & .0100 & .0142 & .0165 \\
Perturbation & 27B without layers 44 and 45 & .0100 & .0142 & .0043 \\
Activation shift & 2B & .0122 & .0131 & .0072 \\
Activation shift & 9B & .0098 & .0050 & .0056 \\
Activation shift & 27B & .0135 & .0106 & .0152 \\
\bottomrule
\end{tabularx}
\end{table}

Sofroniew et al.~measured depth on activations, so we also put the depth question to the activation shifts of \S7.2. At each layer we take the absolute value of each emotion's shift and average over the 171 emotions. This mean shift does not grow steadily with depth. Averaged by thirds, it rises then falls at 2B and falls then rises at 9B and 27B (Table~\ref*{tab:5}). Our activation measure differs from theirs, so this difference between Gemma 2 and Claude cannot be attributed to the models (Appendix G).

Figure 84 of \citet{sofroniew2026emotion} also shows that post-training's early-layer and late-layer shift profiles correlate weakly, appearing as two blocks in a layer-by-layer correlation heatmap. The analogous structure appears at every Gemma 2 size. We correlated the activation shifts between every pair of layers and searched for the single split that best separates early layers from late ones. Layers on the same side of the split correlate at .53 to .65, and layers on opposite sides near zero ($-$.09 to +.14), so tuning's early-layer and late-layer shifts are largely separate changes. Two null analyses indicate that the separation exceeds what the split search alone produces, but they do not establish a reliable boundary layer (Appendix G).

\subsection{The exception: a stable ranking of displaced emotions}

The tuning contrasts of \S7.1 are small in aggregate, and the activation shifts of \S7.2 do not agree across sizes. That leaves open whether tuning moves every emotion a little or a few emotions a lot, and whether the few are the same at every size. We ask with displacement, a registered measure defined per emotion (Appendix C). A tuning pair is a base cell and its tuned cell at one size. For each pair, we align the two cells' leading two axes, as in \S4.2. After the alignment each emotion has one position from each cell, and its displacement is the distance between the two. We rank all 171 emotions by displacement. The ranking, the readings drawn from it, and the regression below are unregistered and exploratory. The registered noise reference for displacement was not measured, and redraws of the cleanup basis test the ranking's stability instead (Appendix G).

Vindictive, vengeful, and spiteful hold the top three places at 27B but not at the smaller sizes (Appendix G). The three guilt-related emotions, regretful, remorseful, and guilty, rank high at every size. Regretful ranks in the top five of 171 in all three tuning pairs, guilty in the top six, and remorseful in the top thirteen. The family that holds compassionate, grateful, kind, and loving in every cell never leads. The rankings agree across sizes, with a Spearman correlation of .81 between 2B and 27B over all 171 emotions (.790 over the 137 the covariate regression uses), where the activation shifts of \S7.2 at the same two sizes correlate at only .23. The activation side agrees with the ranking on the three guilt-related emotions, which are among the ten emotions tuning lowers most at 27B (\S7.2), and not on the three vindictive emotions. Displacement and activation shift are separate computations, so that agreement is not one result counted twice, and the three guilt-related emotions are not merely unstable emotions, since their purity sits above average in four of the six cells (Appendix G).

A plausible explanation is that tuning displaces the self-conscious moral emotions, such as guilt and remorse. We tested it with a regression of displacement on an indicator for eight such emotions fixed in advance (\citealp{haidt2003moral}; \citealp*{tangney2007moral}), seven of which enter the regression (Appendix G), with five covariates, pooled across the three tuning pairs. The indicator shows no detectable effect ($\beta$ = +0.25, p = .81), but only seven emotions carry it, so the test cannot rule out a large effect in either direction. Two covariates do predict displacement, arousal ($\beta$ = +0.24, p = .005) and layer-0 readability, with more readable emotions moving further ($\beta$ = $-$0.25 on log rank, p = .014), yet a model built on them at 2B predicts the 27B ranking poorly (Spearman .292). The ranking is stable across sizes, and the predictors that would explain it are not. That tuning displaces emotions in proportion to arousal and layer-0 readability is a hypothesis this study did not register and a future study could test.

\section{Robustness checks}

The checks in \S4 to \S7 each test one finding against one alternative. This section revisits aspects of the pipeline that bear on every finding at once and merit robustness checks of their own. We rerun the rating-dependent registered analyses of Table~\ref*{tab:E.2} under two further rating sources (\S8.1), and apart from one negative control (\S8.2), no registered verdict changes. Two layer-stability contrasts cleared their floors, and both trace to a single layer of one cell, base 27B's layer 45, whose divergence replicates on re-extraction (\S8.3). The section closes with the adversarial audit that every graded claim passed and with the noise that numerical precision and hardware contribute to the extraction (\S8.4).

\subsection{Other human rating sets leave every verdict but the negative control unchanged}

Every registered correlation with human ratings in this paper follows \citet{sofroniew2026emotion} in using the 45 rated emotions from \citet{russell1977evidence} that overlap with 45 of the 171 emotions with extracted vectors. \citet{hollowell2026replicating} checked base 27B's axes against two further human norm sources, NRC-VAD \citep{mohammad2025nrcvad} and \citet{warriner2013norms}. We rerun the registered analyses that use human ratings against those two sets, in every cell, with the geometry held fixed and only the rating table changed. Three are not rerun: the two pooled tuning off-diagonals and the scale contrast in valence correlation. Both sources enter in two forms, the natural overlap over every emotion the lexicon rates (161 for NRC-VAD, 126 for Warriner) and a matched set of the registered 45 emotions (45 for NRC-VAD and 42 for Warriner, which lacks three of the terms). With Russell and Mehrabian, that makes five human rating settings, and the pass is an unregistered sensitivity analysis.

No registered verdict changes under any of the four additional settings, apart from a single negative control, which we take up in \S8.2. The tuning nulls stay null, and the two arousal measures of \S5.3 keep their direction in every chain under every setting. The pooled valence delta reaches nominal significance under the largest overlap only (n = 161, p = .036), and its registered verdict stays null. Norm choice moves most effect sizes modestly, and the only verdict it moves is the negative control's (\S8.2). The largest moves are in the arousal-tracking rise of \S5.3, which falls below the .15 floor in both chains on the two natural overlaps (base +.059 and +.077, tuned +.131 and +.129) and in the base chain on NRC-VAD matched (+.145). Its verdict, null after Holm, is unchanged.

\begin{table}[htbp]
\small
\caption{\textbf{Every registered verdict that uses human ratings, under five norm settings.} The registered setting is the 45-emotion Russell and Mehrabian overlap. Per-setting numbers are in Appendix E.}
\label{tab:6}
\begin{tabularx}{\linewidth}{@{}>{\hsize=1.755\hsize\raggedright\arraybackslash\hyphenpenalty=10000\exhyphenpenalty=10000}X>{\hsize=0.890\hsize\raggedright\arraybackslash}X>{\hsize=0.773\hsize\raggedright\arraybackslash}X>{\hsize=0.793\hsize\raggedright\arraybackslash}X>{\hsize=0.897\hsize\raggedright\arraybackslash}X>{\hsize=0.891\hsize\raggedright\arraybackslash}X@{}}
\toprule
\textbf{Analysis, with its registered verdict} & \textbf{R\&M, registered (n=45)} & \textbf{NRC-VAD, natural (n=161)} & \textbf{NRC-VAD, matched (n=45)} & \textbf{Warriner, natural (n=126)} & \textbf{Warriner, matched (n=42)} \\
\midrule
Tuning main effect, valence and arousal (null, null) & null, null & null by size, null & null, null & null, null & null, null \\ \addlinespace[3pt]
Arousal tracking with scale (lead, not confirmed) & positive, both chains & positive, both chains & positive, both chains & positive, both chains & positive, both chains \\ \addlinespace[3pt]
Tuning by scale interaction (null) & null, positive direction & null, positive & null, positive & null, positive & null, positive \\ \addlinespace[3pt]
Embedding-table asymmetry, valence moderate and arousal weak (confirmed) & holds & holds & holds & holds & holds \\ \addlinespace[3pt]
Arousal depth-gap with scale (direction observed, null at the bar) & positive, both chains & positive, both chains & positive, both chains & positive, both chains & positive, both chains \\ \addlinespace[3pt]
Negative control, arousal off the second component (0 of 6) & 0 of 6 & 5 of 6 & 0 of 6 & 2 of 6 & 2 of 6 \\
\bottomrule
\end{tabularx}
\end{table}

Of note in Table~\ref*{tab:6} is the embedding-table asymmetry, which we observed in \S5.1. The moderate correlation with valence and the weak correlation with arousal hold across five human rating settings, so the choice of \citet{russell1977evidence} is not what produces it. The depth gap (\S5.3) is measured from the weak embedding-table arousal correlation, and our interpretation of the readability gate in 27B cells (\S6.2) relies on valence being present at the embedding table. The stability of the asymmetry across ratings sets strengthens those findings.

\subsection{The arousal axis depends on the emotion set}

\S8.1 names a single negative control that moves with the rating setting, and it concerns the arousal axis. We outline the control first, then how it moves under the five human rating settings.

The main analysis (\S4.1) fits principal components on all 171 emotions and scores the 45 of them that overlap with \citet{russell1977evidence}. The control fits the components on those 45 emotions alone and asks whether arousal still lands on the second component. We registered the control for arousal because a change of emotion set is more likely to move the weaker of the two axes, and we registered two predictions: that arousal would leave the second component when the components are fit on 45 emotions, and that tuned cells would keep arousal on the second component more often than base cells.

In the control, arousal drops from the second component to the third in all six cells, with best correlation magnitudes of .67 to .77, so each cell still carries an arousal axis, but 0 of 6 carry it as the second component. No tuned cell keeps arousal on the second component (0 of 3), so tuning does not differ from base on this measure.

The second and third components carry nearly equal variance across cells when fit on the 45 emotions, which makes the rank fragile. 0 of 6 here is a point estimate, the single value from one fit, and resampling shows how it changes over repeated fits. In an unregistered check that re-grades nothing, we resampled the 45 rated emotions 1,000 times, with each draw shared across the six cells. Arousal lands on the second component in .25 to .36 of draws per cell. The 0 of 6 is the most common outcome across draws but recurs in only .59 of them, and all six cells land on the second component together in .22 (Appendix E, Table~\ref*{tab:E.3}). With the primary set of 45 emotions, then, the second and third components are close to a tie, and which one arousal falls on changes from fit to fit in each cell. The instability is consistent with \citet{sun2026valence}, who found no single component for arousal on 27 emotions. \citet{hollowell2026replicating} refit base 27B on random subsets of 27 emotions and reproduced their null in most of them.

The additional human ratings sources of \S8.1 let us ask two further questions of this control. First, whether its outcome depends on Russell and Mehrabian's particular ratings. On NRC-VAD's matched set of the same 45 emotions, arousal lands on the second component in 0 of 6 cells, and on Warriner's matched 42 emotions in 2 of 6. Second, we can ask whether the size of the rated set matters. Fitting on 161 emotions for NRC-VAD and 126 for Warriner restores arousal to the second component under NRC-VAD (5 of 6) and not under Warriner (2 of 6), on point estimates and under resampling alike (Table~\ref*{tab:E.3}). In short, this establishes no direction across sources. The two larger sets also differ in composition and coverage at once, and the series cannot separate the two. Under resampling in every setting, the registered one included, arousal's principal component rank changes from draw to draw in every cell.

\subsection{One divergent layer behind two registered verdicts}

\citet{sofroniew2026emotion} measured how steady Claude's emotion geometry stays across fourteen evenly spaced layers, and we registered a measure of that stability for every cell. We sample fourteen layers, which in the 46-layer 27B cells are layers 0, 3, 7, 10, 14, 17, 21, 24, 28, 31, 35, 38, 42, and 45. At each sampled layer we take the 14,535 cosines among the 171 cleaned emotion vectors (\S3), and we correlate those cosine tables between layers. Mid-late layer stability is the average of those correlations among the seven sampled layers at or past half depth, 24 through 45 at 27B, so it is high when the geometry changes little from one late layer to the next. Three registered contrasts test mid-late stability against a floor of .01: tuning, pooled over the three sizes, and scale from 2B to 27B in each chain (Table~\ref*{tab:E.4}). A fourth registered contrast tests tuning on stability averaged over all fourteen layers.

Two of the three mid-late contrasts cleared the floor. Tuning raises stability by +.0142 pooled, and scale lowers it in the base chain by $-$.0424, both real by the case bootstrap fixed at grading (\S3). The scale contrast in the tuned chain is sub-floor ($-$.0003), and the all-layer tuning contrast is sub-floor at +.0007.

A verdict can be real and still rest on a single layer, so we drop layer 45 and regrade. Base 27B's mid-late stability is .949, against .992 to .993 in the other five cells, because its cosine table at layer 45 agrees poorly with its tables at the other six late layers. Dropping layer 45 from both 27B cells lifts base 27B to .994, just above the range of the other five cells, sends the tuning contrast to zero at this precision, and reverses the sign of both scale contrasts (Appendix H). The two contrasts that cleared the floor therefore rest on one fact: base 27B's geometry departs from its own late geometry at layer 45, and tuned 27B's barely does, which is why the tuning contrast is carried by the 27B pair (Table~\ref*{tab:4}).

The divergence belongs to the model and not to one extraction run. We re-extracted base 27B at three of the sampled layers, 45, 42 before it, and the analysis layer 31 as a control, twice with different seeds under a rule fixed before the second seed was read, and the divergence appeared on both seeds. A sweep of every layer from 42 to 45, including the unsampled 43 and 44, shows that it builds gradually, steepest between layers 43 and 44, and spreads across many dimensions, and that the tuned cell's divergence is much milder (Appendix H).

This finding is why we reported the 27B late third with and without its final two layers (\S7.3), since final layers are known to specialize for the output task \citep{voita2019bottomup,lad2024stages}. It is also why we recommend pairing any depth-band contrast with a check that drops one layer at a time. Why base 27B diverges at its final layers is an open question, and whether other families' base models do the same is one a second family grid would answer (\S10).

\subsection{Audit and the cosine noise floor}

Every claim carrying a graded verdict in this study passed an adversarial audit, and the prompts and reports ship as supplementary artifacts. Each audit extended the protocol of \citet{hollowell2026replicating}, where a fresh-context Claude session received the claim and the artifacts but nothing of the session that produced them. We instructed the session to refute the claim by independent recomputation from the artifacts.

Most audits received the claimed values as targets. A smaller set received none, derived every number from the raw inputs, and wrote the audit results down before opening the artifact to compare. A separate fresh-context grader, with no connection to either the producing or the auditing session, proposed each grade, and the final rulings were ours.

The seven audits reproduced every number the paper takes from the audited files, using independent code. We release the audit protocol in the hope that it is reusable.

We read cosine tables to the third decimal and report here which decimals this pipeline can reproduce. The first source of noise is precision. Vectors extracted in bf16 and in fp32 agree at a mean per-story cosine of .999994 or better at all three sizes, so precision adds almost none.

The second source is hardware. Kernel and hardware paths shift numerics through reduction orderings that depend on the batch, and the control that literature recommends is a twin run \citep{golden2024flash,he2025nondeterminism}. Stored extractions at 9B base and 27B base, compared against fresh ones on a later accelerator, move by a few parts in a thousand. On one accelerator, then, the fifth decimal of a per-story vector's cosine is noise, and between two accelerators the fourth is. These bounds are measured on single vectors and are not propagated through the averaged and centered quantities the paper compares. Appendix J reports where the error falls at each size and the twin run that anchors the hardware comparison.

\section{Discussion and limitations}

This section discusses the results of the Gemma 2 family grid (\S9.1) and outlines the study's limitations (\S9.2).

\subsection{Discussion}

\citet{sofroniew2026emotion} found a core emotion geometry in Claude Sonnet 4.5, and \citet{hollowell2026replicating} replicated it in Gemma 2 27B base. We asked whether that emotion geometry holds across the Gemma 2 family, and answer that it does. Every cell exhibits a strong valence axis, a weaker arousal axis, and soft but stable emotion families (\S4.1). The geometry corresponds across cells closely enough that an emotion's vector in one cell predicts its vector in another (\S4.3).

We also asked how emotion geometry changes with scale, depth, and instruction tuning within the Gemma 2 family. We answer that scale spreads the geometry's variance across more components (\S5.2), that at 27B the emotion words stay unreadable through the output vocabulary for half the network (\S6.2), and that tuning leaves the geometry at the analysis layer within every registered floor (\S7.1).

Scale, more specifically, largely preserves the emotion geometry's shape while altering its distribution. From 2B to 27B every cell keeps the same axes and closely matching families, and the planes of the nine compared pairs align almost without rotation (\S4.2). Larger models put less of the emotion variance on the leading component and spread it across more directions (\S5.2). The arousal correlation also rises with size, a lead the registered tests did not confirm (\S5.3). The most significant discontinuities by scale in the paper sit at 27B. The readability gate appears only there (\S6.2), base 27B's final layer departs from its own late geometry where no smaller cell's does (\S8.3), and tuning's activation shifts diverge there (\S7.2). Within this family, 2B to 9B is a change of degree and 9B to 27B is where new behavior appears.

Depth matters more as the model grows. In the 2B and 9B cells the emotion words read from the first layer (\S6.2), and the geometry at the final layer agrees with the geometry at the layers before it (\S8.3), so the emotion geometry of a smaller model is close to the same object at every depth. The 27B cells, in contrast, have phases. For the first half of the network the emotion words cannot be read from the output vocabulary, even though valence is present from the embedding table onward (\S5.1), and at half depth they become readable in a single step (\S6.2, \S6.3). At 27B the geometry holds across the middle layers, yet in the final layers the base model's geometry moves (\S8.3) and tuning's perturbation rises (\S7.3). Depth reveals that the presence of emotion geometry and the readability of emotion words are, in the Gemma 2 family, different phenomena.

Instruction tuning leaves the base model's geometry largely in place. Every base cell exhibits the core emotion geometry, and no registered measure at the analysis layer changes beyond its floor under tuning (\S4.1, \S7.1). This ran against our registered prediction, but accounts of persona selection predict the pattern. For example, the Persona Selection Model \citep*{marks2026persona} reads post-training as eliciting an assistant persona from a repertoire the base model already represents, and a geometry the base model already carries is a repertoire of that kind.

Tuning is not entirely inert, though. Its perturbation of the geometry grows with depth at every size and at 27B comes from the last two layers (\S7.3). Tuning changes what the emotion vectors read from neutral text differently at 27B than at the smaller sizes (\S7.2), and the emotions it displaces most form a stable ranking across sizes (\S7.4). Where tuning has an effect, this study's measurements place it late in the network and in a few emotions rather than in the geometry as a whole.

Across all three axes, the geometry the family varies is one it largely starts with. We read the family's emotion geometry as substantially inherited from lexicon and corpus. Once cleaned of generic token statistics, valence reads off the static token-embedding table in every cell, with no forward pass (\S5.1). The correspondence between models is already present in those tables, since transport maps refit on embedding-table vectors keep median rank 1 in every pair, as maps fit at the analysis layer do (\S4.3, \S5.1). And the six models, across three sizes and both tunings, build nearly the same pairwise similarities, valence-arousal plane, and emotion families on top of that embedding table, within a small margin of the instrument's agreement with itself (\S4.2).

Why does the table carry so much of the emotion geometry and yet so little of arousal? Valence is the geometry's largest component, holding about twice the emotion variance that arousal does in every cell, and two and a half times at 2B (\S4.1), and valence is the axis the table already supplies (\S5.1). Each model's layers add to both axes, but arousal starts from almost nothing at the table and owes nearly all its strength to the forward pass (\S5.1). The arousal axis is also the one that behaves least like a fixed property of the space, since scale may strengthen it (\S5.3) and the emotion sample moves it (\S8.2). One potential explanation, that valence is lexical and arousal is situational, cannot be adjudicated by these tests alone.

How far these results reach past the Gemma 2 family remains open. \S10 sets out this and other open questions that this study could not settle.

\subsection{Limitations}

\begin{itemize}
\item
  Following \citet{sofroniew2026emotion} and \citet{hollowell2026replicating}, we assume throughout that emotion concepts are represented as linear directions in activation space. A single vector may capture blended or complex emotions poorly, and a linear vector is generic with respect to binding and directionality toward characters. Like \citet{sofroniew2026emotion}, we cannot be certain the vectors ``capture all or only the emotion concepts we intend,'' nor that they are free of confounds.
\item
  This study is observational and measures representations of emotion concepts, with no steering and no behavioral measure. No result here shows that or how Gemma 2 models use these representations when they generate text, and neither the presence of emotion geometry nor the readability of emotion words licenses that inference (\citealp{elazar2021amnesic}; \citealp*{ravichander2021probing}).
\item
  The pipeline of \citet{hollowell2026replicating}, which replicated \citet{sofroniew2026emotion}, is inherited unchanged and its limitations carry over. The corpus of emotion stories is generated by \texttt{claude-sonnet-4-5-20250929} and is English only, and its validation rests on one human coder, the author, hand-coding 300 sampled stories with no inter-rater statistic. Every Gemma 2 family cell reads the same 205,200 stories, so the transport result of \S4.3 concerns the emotion centroids of one shared corpus, and whether transport holds for independently written text remains untested here.
\item
  Six cells of Gemma 2 share a pretraining lineage, a tokenizer, and data, so this study does not separate facts about this family from facts about transformers. The three sizes also differ in training regime and not only in parameter count. The Gemma 2 report trains the 2B and 9B checkpoints by distillation from a larger model and the 27B checkpoint from scratch (\hyperlink{cite.gemmateam2024gemma2}{Gemma Team}, \citeyear{gemmateam2024gemma2}), so the size comparisons do not isolate parameter count, and the 27B departures of \S9.1 could reflect training from scratch as well as size. The tuning test likewise covers instruction tuning in one family only.
\item
  Statistical power on every registered contrast against human ratings is bounded by the 45 emotions that Russell and Mehrabian rated, not by the story sample (\S5.3, the \S8 contrasts are unregistered). The .15 floor inherits the Fisher tests' assumption that the two correlations in a contrast come from separate samples, and because both are scored against the same 45 emotions the floor is higher than the data required (\S5.3, \S7.1). The .10 floor on the norm correlations rests on the same assumption, and how far it overstates the requirement is not measured. A result above a floor would clear a correctly sized one, so above-floor verdicts stand, while a sub-floor null is weaker evidence of absence than the floor suggests.
\item
  The adversarial audits (\S8.4) and the analyses they audited ran on the same model family. Their independence is contextual rather than a matter of model.
\item
  Readability results in \S6.2, \S6.3, and Appendix F depend on the logit lens and hold within the grid's own contrast between cells and not against the lens. Lenses degrade away from the final-layer basis \citep{belrose2023tuned}, so ``readable'' in this paper means readable when the emotion vector is projected through the final normalization and unembedding, and a different readout could place the readability gate at a different layer or find it absent altogether.
\end{itemize}

\section{Open problems}

This section points to open problems that our analyses sharpened but could not settle. Each entry indicates what a reliable result would require and the features of a study design that would deliver it. We release the code and artifacts to make several of these runnable without regenerating the full emotion stories corpus, and we invite collaboration with anyone who wants to take one up.

\textbf{What produces the 27B gate.} In both 27B cells the median emotion word stays outside the top 10,000 until layer 23, while in the four smaller cells it reads from layer 0 (\S6.2). The emotion word may already be encoded in the early 27B layers, but along directions the unembedding does not read, so the projection cannot see it. The early layers may not yet encode the word, and the middle layers may build it. The readout itself may contribute, since it projects a difference-of-means vector through an instrument built for residual states (\S6.1), though that mismatch holds in every cell and the gate would then be tied to something specific to 27B's residual stream, such as its massive-activation dimension (Appendix J). The next test is an alternative readout, a tuned lens \citep{belrose2023tuned} fit in every cell on text disjoint from the stories and applied to the story activations before the emotion contrasts are taken, with the own-word trajectory and the seam comparison of \S6.3 rerun under it and reported beside the present readout. If the alternative readout recovers the emotion words well before layer 23, the gate depends on the projection. If it moves the onset relatively little, the gate survives a change of instrument, which does not yet place it in the model's computation but makes the seam the place to test next.

\textbf{The cross-family grid.} Whether the results from the Gemma 2 family hold across other model families remains untested. The nearest cross-family scale series is \citet*{tagliabue2026pain}, whose 25 models include Qwen 2.5 at 7B, 32B, and 72B, base and tuned, though they measure one direction's separation rather than the geometry of a full emotion set. A subsequent study could apply the same grid in a second family with base and tuned checkpoints, with transport maps fit between the two families as well as within the second, so that correspondence across families is tested and not only replicated within one. The second grid would also show whether a different base model's final layers diverge as base 27B's do (\S8.3). Ideally, the second family's training regime should be held fixed across sizes, since Gemma 2 trains its 27B checkpoint from scratch and the smaller two by distillation (\S9.2). At least one pair of cells in each family should read both the original stories and a corpus of emotion stories written by a different generator, so that failures can be assigned to the family or to the corpus rather than both at once. Readability should be compared on emotion words that are single tokens in both tokenizers.

\textbf{What the embedding table carries forward.} PC1 of the vectors drawn from the embedding table correlates with human valence ratings at .48 to .51 in every cell, and PC1 of the vectors at $\sfrac{2}{3}$ depth correlates with them at .71 to .78 (\S5.1). It remains open how much the layers strengthen the valence structure already present in the embedding table and how much they develop a different structure that also tracks human valence ratings. \citet{hollis2016principals} found a component that tracks human valence in the static word vectors of a skip-gram model trained on text alone, so valence structure in a token-embedding table is expected, and the layers' contribution is the informative part. Within each cell, the valence direction at the table and the valence direction at the $\sfrac{2}{3}$-depth analysis layer use the same residual-stream coordinates. Two measurements would quantify their alignment: the cosine between the two directions and the share of emotion variance at the analysis layer that lies along the table's valence direction. Testing whether the later structure depends on the earlier one, rather than merely aligning with it, would require altering activations along the table's valence direction and measuring the effect at later layers, with controls to distinguish a specific effect on valence from general disruption. Both alignment measurements are computable from the released vectors at the table and at the analysis layer.

\textbf{The mechanism of variance spreading.} We found that Gemma 2's larger models put less of their emotion variance on the leading component, while the participation ratio rises with scale and the TwoNN point estimates stay similar (\S5.2). The next test is held-out prediction of emotion identity and of human ratings from the leading components alone, from the remaining components alone, and from matched component budgets across cells, which asks whether variance outside the leading directions contributes signal. If it does, sparse-feature decomposition on the toolchain \citet{presa2026compositional} validated for Gemma 2 9B tuned is the instrument for saying what that signal represents. Their finding that joy and adoration appear only under steering of several features together is one candidate answer.

\textbf{Arousal correlation with scale.} We observed that the arousal correlation on the second component rises from 2B to 27B in both chains, though we classified this as a lead rather than a result because registered tests did not confirm it (\S5.3). Larger models may hold more arousal information, or the second component may simply catch a larger share of the same amount, since the second and third components together recover about as much arousal in every cell (Appendix E) and reading 2B one layer deeper closes most of the base chain's gap (\S5.3). The next test is a measure of the arousal linearly recoverable from a pre-stated low-dimensional subspace, scored on held-out emotions over a depth band rather than at one layer. That measure would split the second-component rise into two parts, the share that is new arousal information and the share that is arousal moving between components.

\section{Reproducibility statement}

The code, result files, and filings are released at \href{https://github.com/adamhollowell/emotion_probes/tree/main/gemma_grid}{\nolinkurl{github.com/adamhollowell/emotion_probes/tree/main/gemma_grid}}, a subdirectory of the replication study's repository, whose own release at commit cdae2fa supplies the pipeline this study inherits unchanged. Large binaries are in the companion artifact repository at \href{https://huggingface.co/adamhollowell/gemma-emotion-vectors}{\nolinkurl{huggingface.co/adamhollowell/gemma-emotion-vectors}}, under gemma\_grid/. The release manifest pins the software environments, Appendix J gives the compute, and Appendix I gives the license terms for every third-party asset. Items that cannot ship as files are released as recipes, with the constraint stated in each case.

\textbf{Vectors and activations.} The release holds six per-cell vector files, each with cleaned and raw vectors at every layer. The base 27B file is the replication study's released vectors. The six embedding-table vector files, cleaned and raw. The six neutral-dialogue activation files, from which the cleanup bases and the cleaned vectors are recomputed. The six cleanup bases themselves. The six per-story activation slices, one per model, each including the model's analysis layer. The per-emotion mean activations from the base 27B re-extractions behind the layer-45 replication and the layer sweep (Appendix H), and the tuned 27B model's colon-token activations (Appendix G). The vector files, the embedding-table files, the neutral-dialogue activation files, the cleanup bases, the per-story slices, the bootstrap draws, and the re-extraction and colon-token activation files are Gemma Model Derivatives, distributed under the Gemma Terms with the notice and use restrictions (Appendix I).

\textbf{Resampling draws and derived tables.} The 1,000-draw story bootstrap, the 1,000 cleanup-basis redraws behind the displacement ranking (\S7.4), and the 500-draw seam and onset bootstrap (\S6.3), each with its per-draw values. The 2,000-draw case bootstraps ship as summaries, and each producer is seeded to regenerate its draws. The per-emotion displacement tables for the three tuning and six scale pairs, and the per-cell family membership tables. No fitted transport map is saved, and the code refits every map from the released vectors.

\textbf{Code.} Every analysis script and figure producer, keyed in the README to the result it produces. The released scripts carry their release names, and the README maps each to the name it ran under. Where a released script differs from the copy that ran, a record linked from the release manifest gives both hashes and the changed lines. The scripts whose hashes the paper's records cite also ship as run. The README also identifies three producers we rewrote after the fact from the procedure their artifacts record. Two more scripts come from the surviving record of the run that used the transport driver: the original driver, which we recovered, and a new script that rebuilds the inputs read by the cross-cell transport (\S4.3), the displacement regression (\S7.4), and the frequency control (Appendix F). The rewritten producers and the transport driver reproduce their stored outputs exactly, and those three analyses do so from the rebuilt inputs.

\textbf{Registration and filings.} The preregistration as last saved, with its hash. The eight pre-stated filings, unedited. Six are fixed by their hashes. Two are fixed by their dates, the filing for the layer-45 replication rule (Appendix H) and the filing for the colon-token design (Appendix G). The dated amendment to the disjoint-topic filing, filed before the part it withdraws was run. The registration concordance.

\textbf{Audit record.} The seven audit prompts as issued and reports as returned, with the fresh-context grading of \S6.1 as a 72-row verdict file linked by hash to the full grading run.

\textbf{Carried in the replication study's repository.} The 45-emotion Russell and Mehrabian table with its provenance. The 500 neutral dialogues, at commit cdae2fa. The base 27B model's scenario activations, which the colon-token comparison reads (Appendix G).

\textbf{Corpus and external inputs.} The 205,200-story corpus is not redistributed by either release. Its generation code and the verbatim prompt and topic design are in the replication study's repository and that paper's Appendix A. The release ships the emotion and topic of every story, with no story text, and the map from slice rows to stories, so the disjoint-topic fits rerun from the slices. The bag-of-words baseline of the disjoint-topic analysis (\S4.3, Appendix D) reads the story text, so it does not reproduce from the release alone. The Claude comparison in Appendix G reads Table 16 of \citet{sofroniew2026emotion}, which neither release redistributes. NRC-VAD and \citet{warriner2013norms} are third-party and are fetched by script under their own terms.

\textbf{Released as an export recipe.} The vocabulary-readout ingredients are not bundled. The unembedding matrix and final-norm weight of each of the six checkpoints are Gemma weights, so the release ships the script that exports them. A user regenerates the tensors from the six checkpoints and rebuilds the vocabulary map from each checkpoint's tokenizer. The export reproduces every stored tensor exactly, and the vocabulary map rebuilt from each checkpoint's tokenizer matches the stored map in all 256,000 entries.

\section*{Acknowledgements}
\addcontentsline{toc}{section}{Acknowledgements}

This is an independent study. It was not conducted in collaboration with, funded by, reviewed by, or endorsed by Anthropic or Google. Claude is a trademark of Anthropic, PBC. Gemma is a trademark of Google LLC. Model names are used solely to identify the systems studied.

\bibliographystyle{plainnat}
\begingroup\raggedright  
\bibliography{refs}
\endgroup

\clearpage 
\appendix
\makeatletter
\@addtoreset{table}{section}
\@addtoreset{figure}{section}
\makeatother
\setcounter{table}{0}
\setcounter{figure}{0}
\renewcommand{\thetable}{\Alph{section}.\arabic{table}}
\renewcommand{\thefigure}{\Alph{section}.\arabic{figure}}

\FloatBarrier
\section{Design-space comparison with adjacent studies}

Companion to the sections that engage each neighbor: Zhang and Zhong at \S2, \S5.2, Appendix H and below, Presa and Teles de Oliveira at \S2, \S10 and below, Venkatesh at Appendix F, \citet{han2026welfare} by citation only, and van der Ben et al.~at \S2. This table says what each one measured, so that a reader can check the scoping rather than take it on trust.

\begingroup
\footnotesize
\setlength{\tabcolsep}{4pt}
\begin{xltabular}{\linewidth}{@{}>{\hsize=1.024\hsize\raggedright\arraybackslash\hyphenpenalty=10000\exhyphenpenalty=10000}X>{\hsize=1.021\hsize\raggedright\arraybackslash}X>{\hsize=0.920\hsize\raggedright\arraybackslash}X>{\hsize=0.970\hsize\raggedright\arraybackslash}X>{\hsize=0.911\hsize\raggedright\arraybackslash}X>{\hsize=1.128\hsize\raggedright\arraybackslash}X>{\hsize=1.025\hsize\raggedright\arraybackslash}X@{}}
\caption{\textbf{Design space, this study and its nearest neighbors.}}\label{tab:A.1}\\
\toprule
 & \textbf{This study} & \textbf{\citet{zhang2025decoding}} & \textbf{\citet{presa2026compositional}} & \textbf{\citet{venkatesh2026negative}} & \textbf{\citet{han2026welfare}} & \textbf{\citet{vanderben2026}} \\
\midrule
\endfirsthead
\multicolumn{7}{@{}l}{\textbf{Table~\thetable} (continued)}\\
\toprule
 & \textbf{This study} & \textbf{\citet{zhang2025decoding}} & \textbf{\citet{presa2026compositional}} & \textbf{\citet{venkatesh2026negative}} & \textbf{\citet{han2026welfare}} & \textbf{\citet{vanderben2026}} \\
\midrule
\endhead
\bottomrule
\endlastfoot
Subject models & Gemma 2 at 2B, 9B, 27B, base and instruction-tuned (six cells) & Qwen3 at 0.6B, 1.7B, 4B, 8B and Llama-3 at 1B, 3B, 8B, pretrained and finetuned & Llama 3.1 8B-Instruct and Gemma 2 9B-IT & Llama-3.2 1B, Qwen2.5 1.5B and 3B, all instruction-tuned & Qwen3 4B base and instruct, Qwen3 8B, GPT-OSS 20B & Apertus-8B and Gemma-4-E4B, both instruction-tuned, one size each across two families \\ \addlinespace[3pt]
Stimuli & 205,200 stories across 171 emotions plus 500 neutral dialogues, identical raw text to every cell & roughly 400,000 Reddit-derived utterances, relabeled and synthesized by a single 32B model that also judges & steering prompts under a five-model judge panel, 2 to 7 seeds per cell & 100 clean and corrupted prompt pairs per condition across three domains, sharing one neutral baseline & reinforcement-learning trajectories in a text maze, rewarded against punished & emotion stories against neutral stories, each corpus generated by the model under test \\ \addlinespace[3pt]
Emotion set & 171 & Ekman six plus neutral & 12 base emotions, plus a 27-emotion catalog & valence sign, positive against negative & 171, the same vocabulary, via the concurrent extraction & 171, the same set inherited from \citet{sofroniew2026emotion} \\ \addlinespace[3pt]
Instrument & unsupervised difference-of-means geometry, with principal components, similarity structure, orthogonal alignment, and clustering & supervised probes, a two-layer classifier per layer on frozen states & sparse-autoencoder features at one mid-depth layer, with causal steering & residual-stream activation patching with a logit-gap metric, plus steering & difference-in-means concept vectors from trained activations, read out through the vocabulary & unsupervised difference-of-means geometry with principal components, the nearest match to this study's \\ \addlinespace[3pt]
Human-rating benchmark & \citet{russell1977evidence}, with NRC-VAD and Warriner robustness & none & none, and the judge panel is uncalibrated against humans, which the authors name as their first open problem & none, anchor tokens rather than norms & none, model-judge evaluations & NRC-VAD \\ \addlinespace[3pt]
Axes varied & scale and tuning, within one family & scale and tuning, across two families, probe-side & architecture, tuned models only & none, tuned models only & reinforcement-learning training, with base and instruct variants alongside & none systematic, two instruction-tuned models of similar size across families, varying the generation corpus \\ \addlinespace[3pt]
Depth resolution & every layer & five depths & one layer & every layer & one layer, chosen by maximum linear separability & every layer \\
\end{xltabular}
\endgroup

Table~\ref*{tab:A.1} distinguishes overlap in experimental design from overlap in method. Zhang and Zhong vary scale and tuning, as this study does, but measure supervised probe performance (below). Van der Ben et al.~use difference-of-means vectors, the same 171 emotions, and a human-rating benchmark, but do not systematically vary scale or tuning and include no base model. None of the comparison studies combines a base model, an unrestricted vocabulary readout, and a human-rating benchmark. Han et al.~and van der Ben et al.~also use 171 emotions, while the other comparison studies use smaller emotion sets. This study tests how changing the emotion set affects the recovered arousal axis (\S8.2).

\citet{zhang2025decoding} also report that emotion probes perform at chance at what they call layer 0, the input token embedding before any transformer block, read at the last token alone. Their layer 0 is this paper's embedding table (\S3), not this paper's layer 0. Their result is compatible with the embedding-table result of \S5.1, because that analysis averages the embedding table's vectors over every token from the 50th onward, and an average over many tokens pools the valence each token's embedding carries, where one token's vector carries only its own.

\citet{zhang2025decoding} find that their seven-class probes, six emotions plus neutral, separate emotions better in larger Qwen3 and Llama-3 models. That finding led us to ask whether separability holds in this family as PC1's share falls (\S5.2). The check is not a replication of theirs. We ran an unregistered nearest-class-mean classifier on held-out stories from our corpus, choosing among all 171 emotions rather than seven classes. It picks the right emotion about equally well at every scale, changing by +1.10 points from 2B to 27B in the base chain and $-$0.33 in the tuned chain, while PC1's share falls in the same models. Separability holds in this family as the variance spreads, and the task difference means the result neither matches nor contradicts theirs.

\citet{presa2026compositional} apply the vocabulary projection of \S6.1 to instruction-tuned Gemma 2 9B and Llama 3.1 8B, projecting learned sparse-autoencoder features where this study projects difference-of-means vectors.

\FloatBarrier
\section{Correspondence with the replication study}

Companion to \S4 through \S8. Every analysis this study inherits, where it lives in \citet{hollowell2026replicating}, where it lives here, and what the family adds. Provenance lives in this table so the body sections do not carry it.

\begingroup
\small
\begin{xltabular}{\linewidth}{@{}>{\hsize=0.945\hsize\raggedright\arraybackslash\hyphenpenalty=10000\exhyphenpenalty=10000}X>{\hsize=0.923\hsize\raggedright\arraybackslash}X>{\hsize=0.781\hsize\raggedright\arraybackslash}X>{\hsize=1.351\hsize\raggedright\arraybackslash}X@{}}
\caption{\textbf{Inherited analyses and their family extensions.}}\label{tab:B.1}\\
\toprule
\textbf{Analysis} & \textbf{In \citet{hollowell2026replicating}} & \textbf{Here} & \textbf{What the family adds} \\
\midrule
\endfirsthead
\multicolumn{4}{@{}l}{\textbf{Table~\thetable} (continued)}\\
\toprule
\textbf{Analysis} & \textbf{In \citet{hollowell2026replicating}} & \textbf{Here} & \textbf{What the family adds} \\
\midrule
\endhead
\bottomrule
\endlastfoot
Core geometry, components and axes & \S5.1, \S5.2 & \S4.1, Table~\ref*{tab:1} & Six cells rather than one, with the base 27B row reproducing the published values \\ \addlinespace[3pt]
Human-norm alignment & \S5.2 & \S4.1, \S8.1 & Same 45-emotion benchmark at every size and tuning level, plus two further norm sources \\ \addlinespace[3pt]
Emotion-set control & \S9.3 & \S8.2 & The concurrent-work reconciliation extended from one model to six, with comparisons across three norm sources and matched subsets \\ \addlinespace[3pt]
Static token-embedding table & \S6.6 & \S5.1 & Moderate valence correlations and weak arousal correlations, significant in no cell, shown in all six cells \\ \addlinespace[3pt]
Within-model arousal measurement & \S6.4 & \S5.3 & Run in six cells, where the joint second-and-third correlation is flat and the second component's own share of it grows with size in both chains at the analysis layer, with the base chain's rise depending on 2B's layer (\S5.3) \\ \addlinespace[3pt]
Consolidation seam & \S6.2 & \S6.3 & The seam shown to move with scale, from quarter depth to half depth \\ \addlinespace[3pt]
Cross-layer stability, late band & \S6.5, Table 2 & \S8.3, Appendix H & Base 27B's late-layer divergence independently replicated under a pre-stated rule, and measured layer by layer \\ \addlinespace[3pt]
Top-token projection & \S7.1 & \S6.1 & The same operation in six cells, graded by a fresh-context session against a rubric fixed before the data \\ \addlinespace[3pt]
Readability onset & \S7.3 & \S6.2 & The onset shown to be a 27B-only phenomenon, absent in every smaller cell \\ \addlinespace[3pt]
Normalized-lens readout & \S7.2 & \S6.2, Appendix F & Applied family-wide, and split by valence sign \\ \addlinespace[3pt]
Residual-norm growth & \S6.2 & Appendix H & Family-wide curve, with the late transition localized to the 27B cells \\ \addlinespace[3pt]
Layer-to-layer rotation of emotion vectors & Released analysis accompanying \citet{hollowell2026replicating}, not reported in its text & Appendix H & Family-wide, with the late-band rotation regime localized to base 27B \\ \addlinespace[3pt]
Valence asymmetry in readability & \S10.1 & Appendix F & Base models included, frequency controlled \\
\end{xltabular}
\endgroup

\FloatBarrier
\section{Registration concordance}

Companion to \S1's definition of the registered level and to \S3. The preregistration was locked on 2026-07-20, before the five new cells were extracted and with base 27B's published values in hand, and is released with this study. It fixed the signals, their predicted directions, the floors, the test for each kind of measure, and the correction scheme. The rows in each correction family, the endpoint form of the scale contrast, the self-agreement test, and the resampling unit of the layer-stability bootstrap were fixed in the grading code on 25 July 2026, after extraction and before any contrast was graded, with each cell's own measured values in hand, so those choices carry the pre-stated level of commitment and not the registered one. Four statuses are used. As registered: tested or reported as the registration specified. Operationalized differently: tested, with a named departure from the registration's wording. Reported without a grade: a registered quantity reported descriptively because the registration gave it neither a test nor a floor. Not reported: a registered quantity or directional prediction that this paper does not report. Table~\ref*{tab:C.1} gives every directional prediction and its outcome. Table~\ref*{tab:C.2} gives every registered item whose reporting departs from, qualifies, or omits the registration. The full concordance, with every registered item and its status, is released with this study beside the registration. Pre-stated analyses fixed after the registration (the transport battery, the \S6.3 calibration, the Appendix G prediction, the Appendix H replication rule) are not in these tables and are disclosed where they appear. Registered items concerning the loving vector belong to a separate study, not yet released, and are not listed.

\begingroup
\small
\begin{xltabular}{\linewidth}{@{}>{\hsize=0.980\hsize\raggedright\arraybackslash\hyphenpenalty=10000\exhyphenpenalty=10000}X>{\hsize=1.163\hsize\raggedright\arraybackslash}X>{\hsize=1.341\hsize\raggedright\arraybackslash}X>{\hsize=0.570\hsize\raggedright\arraybackslash}X>{\hsize=0.946\hsize\raggedright\arraybackslash}X@{}}
\caption{\textbf{Directional predictions and their outcomes.}}\label{tab:C.1}\\
\toprule
\textbf{Prediction} & \textbf{Scale (2B to 27B)} & \textbf{Outcome} & \textbf{Tuning (base to tuned)} & \textbf{Outcome} \\
\midrule
\endfirsthead
\multicolumn{5}{@{}l}{\textbf{Table~\thetable} (continued)}\\
\toprule
\textbf{Prediction} & \textbf{Scale (2B to 27B)} & \textbf{Outcome} & \textbf{Tuning (base to tuned)} & \textbf{Outcome} \\
\midrule
\endhead
\bottomrule
\endlastfoot
1 Variance shares (M) & PC1 share up modestly; PC2 to PC3 gap widens & PC1 fell by about five points in both chains, graded real in the opposite direction (\S5.2, Table~\ref*{tab:E.4}). The PC2 to PC3 gap was never operationalized as a statistic and is ungraded (Table~\ref*{tab:E.4}). PC2's share alone was graded, null by floor & PC1 and PC2 up slightly & PC1 +0.04 points, null. PC2 fell 0.32 points, reliably nonzero, null by floor. PC3 rose 0.15, null by floor (Table~\ref*{tab:4}) \\ \addlinespace[3pt]
2 Valence correlation (H) & Up & +.01 in both chains, null (Table~\ref*{tab:E.4}) & Up & +.056 pooled, null (Table~\ref*{tab:4}) \\ \addlinespace[3pt]
3 Arousal correlation (H) & Up strongly, possibly weak or absent at 2B & +.17 and +.22, above the .15 floor, not significant after Holm, carried as a lead (\S5.3). Arousal is present at 2B (Table~\ref*{tab:1}) & Up & $-$.083 pooled, wrong direction, null (Table~\ref*{tab:4}) \\ \addlinespace[3pt]
4 Off-diagonals (M) & Shrink toward 0 & Not graded on the scale axis. The scale family carries five measures per chain and the off-diagonals are not among them & Shrink & +.041 and $-$.088 pooled, null (Table~\ref*{tab:4}) \\ \addlinespace[3pt]
5 Negative control (M) & Fails at all sizes unless scale strengthens arousal & Fails in all six cells at the point estimate, as predicted (\S8.2, Table~\ref*{tab:E.3}) & Tuned cells more likely to survive & None survives at the point estimate, 0 of 3, opposite to the prediction (\S8.2, Table~\ref*{tab:E.3}) \\ \addlinespace[3pt]
6 Vector norms (expl) & About stable, slight rise & Not reported & About stable & Norm against PC1 position +.095 pooled, under the .10 floor. Norm against absolute PC2 position +.045 (Appendix E) \\ \addlinespace[3pt]
7 Neighbor cosines (M) & High everywhere, slight rise & Not graded on the scale axis & Minor & +.0024 pooled, null (Table~\ref*{tab:4}) \\ \addlinespace[3pt]
8 Self-agreement (M) & Up & Not graded on the scale axis. Reported as a range, .656 to .717 (\S4.1) & Up & $-$.017 pooled, null by the seed-spread test fixed at grading (Table~\ref*{tab:4}) \\ \addlinespace[3pt]
9 Cohesion (M) & Up, the family holding compassionate, grateful, kind, and loving tightens most & Not graded. Reported as a range, .706 to .758 (\S4.1). The prediction for that family is not reported & Up, that family most & +.017 pooled, no floor, reported without a grade (Appendix E). The prediction for that family is not reported \\ \addlinespace[3pt]
10 Cross-layer similarity (M) & Stable high-similarity band at each cell's layer, shorter at 2B & The per-cell band is not reported as such. The scale contrasts on the mid-late band are reported: base chain $-$.042, above the .01 floor, graded real and carried by layer 45 of base 27B; tuned chain $-$.0003, null (\S8.3, Table~\ref*{tab:E.4}) & Minor & +.0142 pooled, above its floor, carried by the same layer (\S8.3) \\ \addlinespace[3pt]
11 Neighbor retention (M) & Stable band at the derived layer & Not reported & Minor & Not reported \\ \addlinespace[3pt]
12 Embedding split (M) & Valence present at the table everywhere, arousal built by depth, the depth gap larger in bigger models & Valence present and arousal weak at the table in every cell (\S5.1). The depth gap grows by +.20 and +.23 in z, registered test not significant, no floor (\S5.3) &  &  \\ \addlinespace[3pt]
i Cross-cell similarity (M) & Tuning pairs agree more closely than scale pairs & Not supported. Tuning pairs did not agree more closely on any of the three levels (\S4.2) &  &  \\ \addlinespace[3pt]
ii Procrustes (M) & Valence axis aligns well everywhere, arousal axis worse, residual larger for scale than tuning & The residual prediction is not supported: the largest and smallest disparities are both scale pairs (\S4.2). The two axes are aligned jointly as a plane, so the axis-by-axis prediction is not reported separately &  &  \\ \addlinespace[3pt]
iii Cross-cell ARI (M) & Scale pairs fall below the .51 floor & Not supported. All nine comparisons clear .51 (\S4.2) &  &  \\ \addlinespace[3pt]
Interaction, arousal (M) & Tuning raises arousal correlation more at 27B than at 2B & +.047, under the .15 floor, p~=~.94, null (\S7.1) &  &  \\ \addlinespace[3pt]
Interaction, cluster tightening (expl) & Roughly scale-independent & Not reported &  &  \\
\end{xltabular}
\endgroup

Confidence is the registration's: H strong hypothesis with prior evidence, M plausible, expl exploratory and outside the Holm family.

\begingroup
\footnotesize
\setlength{\tabcolsep}{4pt}
\begin{xltabular}{\linewidth}{@{}>{\hsize=0.775\hsize\raggedright\arraybackslash\hyphenpenalty=10000\exhyphenpenalty=10000}X>{\hsize=1.082\hsize\raggedright\arraybackslash}X>{\hsize=0.729\hsize\raggedright\arraybackslash}X>{\hsize=1.414\hsize\raggedright\arraybackslash}X@{}}
\caption{\textbf{Registered items whose reporting departs from, qualifies, or omits the registration.}}\label{tab:C.2}\\
\toprule
\textbf{Registration item} & \textbf{The registration says} & \textbf{Reported at} & \textbf{Status} \\
\midrule
\endfirsthead
\multicolumn{4}{@{}l}{\textbf{Table~\thetable} (continued)}\\
\toprule
\textbf{Registration item} & \textbf{The registration says} & \textbf{Reported at} & \textbf{Status} \\
\midrule
\endhead
\bottomrule
\endlastfoot
2B sensitivity & L18 reported as a sensitivity check, ``and a small $\pm$band'' & \S3, \S5.3 & L18 as registered. The band beyond L18 is not reported \\ \addlinespace[3pt]
C1-1 Variance shares & PC1, PC2, PC3 share per cell & Table~\ref*{tab:1} (PC1, PC2), Table~\ref*{tab:4} and Table~\ref*{tab:E.4} (contrasts) & PC1 and PC2 as registered. PC3's per-cell share is not tabulated, only its tuning contrast \\ \addlinespace[3pt]
C1-4 Off-diagonals and permutation p & PC1 against arousal ratings, PC2 against valence ratings, 5,000-shuffle permutation p & \S4.1 (permutation p for both axes), Table~\ref*{tab:4} (tuning contrasts) & As registered for tuning. The off-diagonals' per-cell values are not tabulated \\ \addlinespace[3pt]
C1-6 Vector norms & Norm against PC1 position, norm against absolute PC2 position, norm variability across depth & \S7.1, Appendix E & Tuning contrasts as registered, outside the Holm family as registered. Norm variability across depth is not reported \\ \addlinespace[3pt]
C1-8 Families and self-agreement & k = 10 partition, self-agreement (adjusted Rand index across seeds) & Table~\ref*{tab:1}, Table~\ref*{tab:4}, and the released membership tables & Per-cell values as registered. The tuning contrast on self-agreement has no registered test. Its test, from the spread across seeds, was fixed at grading (\S3, \S7.1) \\ \addlinespace[3pt]
C1-9 Cohesion and boundary & Ten-nearest-neighbor in-family fraction, boundary emotions & Table~\ref*{tab:1}, \S7.1, \S7.4, Appendix E & Reported without a grade. The registration gave cohesion a direction and no floor or test (\S7.1) \\ \addlinespace[3pt]
C2-10 Cross-layer similarity & Cosine of each layer's 171-by-171 cosine table against the analysis layer's & \S8.3 & Operationalized differently: the mean pairwise agreement among fourteen sampled layers, in Pearson, recomputed in cosine with no verdict change. The pairwise form was in the code that set the .01 floor on the day of registration. Appendix H discloses both departures \\ \addlinespace[3pt]
C2-11 Cross-layer neighbor retention & Top-5 retention and top-10 overlap against the analysis layer's neighborhood, all layers & none & Computed in a variant form (a top-5 shared count and a top-10 Jaccard, the registration's assignment reversed) and stored in the released per-cell artifacts. Not graded or reported \\ \addlinespace[3pt]
C3-12 Embedding-table baseline & Valence and arousal correlation, PC variance, similarity structure, ARI against the locked partition, neighbor cosines, all at the static table with the cleanup refit on neutral embeddings & \S5.1, Table~\ref*{tab:E.1}, \S5.3 & Correlations as registered. PC variance, similarity structure, ARI, and neighbor cosines at the table are not reported \\ \addlinespace[3pt]
C4-i Cross-cell similarity & Cosine of two cells' 171-by-171 tables, noise reference measured at the first contrast & \S4.2 & As registered. The within-cell agreement reference was measured on the day the cross-cell point estimates were first computed, after those estimates were written and before any contrast was graded \\ \addlinespace[3pt]
C4-ii Procrustes alignment & Rotation and residual on per-emotion component coordinates, noise reference measured at the first contrast & \S4.2 & As registered, same ordering as C4-i \\ \addlinespace[3pt]
C4-iii Cross-cell partition agreement & Adjusted Rand index between cells' ten-family partitions, with a reorganization floor of .51 set two standard deviations below the mean of a twenty-seed sweep on base 27B & \S4.2 & As registered, and the floor was not revised. Two standard deviations below base 27B's self-agreement in Table~\ref*{tab:1}, .695, would give .54. The lowest cross-cell comparison, 2B base against 27B base at .53, clears the registered .51 and would fall below a floor recomputed from Table~\ref*{tab:1} \\ \addlinespace[3pt]
C4-iv Per-emotion displacement & Post-Procrustes distance per emotion between cells, noise reference measured at the first contrast & \S7.4 & Displacement is a registered signal. The ranking, the readings drawn from it, and the regression in \S7.4 are unregistered and labeled so. The registered noise reference was not measured, and \S7.4 says so \\ \addlinespace[3pt]
C5 Held-out-text activation & Defer to a follow-up & none & Deferred as registered and not run. \S7.2's neutral dialogues and Appendix G's twelve scenarios are separate items and are labeled as such \\ \addlinespace[3pt]
Scale main effect & ``Trend across \{2B, 9B, 27B\} within a fixed tuning level'' & \S5.2, Table~\ref*{tab:E.4} & Operationalized differently: the 2B to 27B endpoint contrast within each chain, fixed in the grading code of 25 July 2026 before any contrast was graded (\S3, \S5.2) \\ \addlinespace[3pt]
Correction families & Holm within three families, tuning, scale, interaction & \S3 & Holm within three families as registered. The registration does not list the rows of each family. The lists (eleven, ten, and one) were fixed in the grading code of 25 July 2026 (\S3) \\ \addlinespace[3pt]
Noise reference for i, ii, iv & Measured at the first contrast and applied before any result is read & \S4.2 & Measured for i and ii from replicate pairs, before grading (see C4-i). Not measured for iv (\S7.4) \\ \addlinespace[3pt]
ARI test & Read against the self-agreement band & \S4.2 & As registered for cross-cell ARI. No test is registered for a within-cell change in self-agreement (see C1-8) \\ \addlinespace[3pt]
Layer-stability test & Reuse the \S6.4 bootstrap code, cross-cell & \S8.3 & Operationalized differently: the replication study's case bootstrap, ported with the resampling unit changed from the 45 rated emotions to the 171, resampling cells in pairs, 2,000 draws, fixed at grading (\S3) \\ \addlinespace[3pt]
Dimension safety & Between-cell signals use only quantities that do not depend on a cell's width, such as cosines and agreement indices & \S4.2, \S4.3 & As registered for the registered comparisons. The transport maps of \S4.3 compare vectors across widths, were pre-stated later, and sit outside this constraint, as \S4.3 says \\
\end{xltabular}
\endgroup

\FloatBarrier
\section{The transport battery in full}

Companion to \S4.3, which summarizes the eight items and the disjoint-topic analysis that followed them. All eight were locked as a single battery before execution and run in one pass.

\begin{table}[H]
\small
\caption{\textbf{Transport robustness, eight pre-stated items.}}
\label{tab:D.1}
\begin{tabularx}{\linewidth}{@{}>{\hsize=0.581\hsize\raggedright\arraybackslash\hyphenpenalty=10000\exhyphenpenalty=10000}X>{\hsize=0.903\hsize\raggedright\arraybackslash}X>{\hsize=1.517\hsize\raggedright\arraybackslash}X@{}}
\toprule
\textbf{Item} & \textbf{Question} & \textbf{Result} \\
\midrule
1. Layer-wise & Does the graded result depend on the analysis layer? & The pinned configuration reproduces the graded object exactly, at median cosine .972. Layer 0 transports as well, at median rank 1 and top-1 .971, so the analysis layer is not privileged over shallow stimulus-driven structure \\ \addlinespace[3pt]
2. Raw vectors & Is the cleanup carrying the transport? & Raw transport matches cleaned, top-1 .994 against 1.0. The confound closes \\ \addlinespace[3pt]
3. Split-half reference & How does cross-cell transport compare with re-measuring one model from half its own data? & Within-model split-half retrieval runs top-1 .947 to .971 and median cosine .944 to .956 across the six cells, with 27B base lowest on median cosine. Cross-cell transport runs higher, at top-1 .994 to 1.0 and primary-pair cosine .972. The split-half figure is computed from half the stories per side and is depressed by that halved sample (\S4.3), so it is a same-construct reference and not a ceiling that bounds the cross-cell result from above \\ \addlinespace[3pt]
4. Valence-arousal ablation & Is the transported object just the circumplex? & With each cell's own leading plane projected out, median rank 1 and top-1 .965. The structure exceeds the plane \\ \addlinespace[3pt]
5. Component truncation & Is the signal carried by a few components? & Retrieval degrades smoothly, from rank 17 at two components to rank 1 and top-1 .743 at ten, and .988 at forty. The signal is distributed \\ \addlinespace[3pt]
6. Centering & Does train-only centering change the result? & No change in median rank, median cosine .972 against .971. The leakage question is retired \\ \addlinespace[3pt]
7. Symmetry & Is transport direction-dependent? & All thirty ordered pairs computed. The fifteen matched directional gaps are near zero, cosine deltas at most .005 \\ \addlinespace[3pt]
8. Family-out & Does the map extrapolate to an unoccupied region? & Leave-one-family-out on the base 27B partition still gives pooled median rank 1 and top-1 1.0. Exploratory and ungraded under the battery's pre-statement \\
\bottomrule
\end{tabularx}
\end{table}

The wording of each verdict in the table is the wording committed when the battery was locked, except item 3, whose committed ranges were found at verification to have truncated both minima by missing the 27B cells. The corrected ranges and the corrected framing are printed, and the committed wording is retained in the released audit record.

One implementation note ships with the battery. In two of 108 decompositions, both at the highest-dimensional pair, 27B base to 27B tuned, and at truncation k = 40, the locked estimator's default solver failed in the environment that ran the battery and an equivalent solver was substituted under a recorded amendment, with every output identical bit for bit. The substitution is driver-level, and the outputs are unchanged.

A pre-stated exploratory analysis, filed after the battery, tests transport when source and target vectors come from disjoint story topics. Under one fixed partition of the 100 topics into two sets of 50, all 32 cross-cell fits meet the median-rank criterion, with top-1 retrieval of .871 to .918 and median cosine of .927 to .940. Cross-cell fits on matched random halves give top-1 retrieval of .924 to .965, while within-cell topic splits give .877 to .924. These reference comparisons are descriptive. One partition does not test robustness across partitions.

\FloatBarrier
\section{Robustness checks on vector cleaning, norm source, tuning, and readability, full tables}

Companion to \S5, \S6, \S7.1, \S8.1, and \S8.2. Table~\ref*{tab:E.1} supports \S5.1 and compares raw and cleaned embedding-table vectors in each cell, with the registered 45-emotion rating set held fixed. Table~\ref*{tab:E.2} supports \S8.1 and gives the rating-dependent registered contrasts we reran under the five norm settings, with the geometry held fixed and only the rating table changed, and Table~\ref*{tab:E.3} supports \S8.2 and gives the negative control under the same settings. Table~\ref*{tab:E.4} is the scale battery in full, and Table~\ref*{tab:E.5} the dependence-aware bootstrap of the tuning battery's correlation contrasts. After the tables, five blocks follow. The first supports \S7.1 and holds the tuning battery's three measures outside the Holm correction, and the other four support \S6.2 and \S6.3 and hold the readability instrument's formula and the checks whose verdicts those subsections report. The appendix holds registered results, pre-stated checks, and unregistered sensitivity analyses, and each table's note states its status.

\begin{table}[H]
\small
\caption{\textbf{Embedding-table correlations with human ratings, raw and cleaned vectors, in each cell.}}
\label{tab:E.1}
\begin{tabularx}{\linewidth}{@{}>{\hsize=1.097\hsize\raggedright\arraybackslash\hyphenpenalty=10000\exhyphenpenalty=10000}X>{\hsize=0.972\hsize\centering\arraybackslash}X>{\hsize=0.979\hsize\centering\arraybackslash}X>{\hsize=0.972\hsize\centering\arraybackslash}X>{\hsize=0.979\hsize\centering\arraybackslash}X@{}}
\toprule
\textbf{Cell} & \textbf{valence r, raw} & \textbf{valence r, cleaned} & \textbf{arousal r, raw} & \textbf{arousal r, cleaned} \\
\midrule
2B base & +.030 & +.485 & +.195 & +.125 \\
2B tuned & +.029 & +.485 & +.193 & +.128 \\
9B base & +.063 & +.507 & +.315 & +.150 \\
9B tuned & +.074 & +.500 & +.325 & +.174 \\
27B base & +.378 & +.479 & +.282 & +.179 \\
27B tuned & +.380 & +.479 & +.288 & +.178 \\
\bottomrule
\end{tabularx}
\end{table}

Each valence value is the Pearson correlation between the emotions' PC1 scores and the \citet{russell1977evidence} valence ratings, and each arousal value is the correlation between their PC2 scores and the arousal ratings, over the 45-emotion overlap, with PC1 oriented to valence and PC2 to arousal. Cleaned vectors have the top three principal components of the neutral dialogues' embeddings projected out (\S5.1). The cleaned columns are the registered embedding baseline, and the raw columns are an unregistered comparison.

\begin{table}[H]
\footnotesize
\setlength{\tabcolsep}{4pt}
\caption{\textbf{Rating-dependent registered contrasts rerun under five norm settings.}}
\label{tab:E.2}
\begin{tabularx}{\linewidth}{@{}>{\hsize=1.525\hsize\raggedright\arraybackslash\hyphenpenalty=10000\exhyphenpenalty=10000}X>{\hsize=0.895\hsize\centering\arraybackslash}X>{\hsize=0.895\hsize\centering\arraybackslash}X>{\hsize=0.895\hsize\centering\arraybackslash}X>{\hsize=0.895\hsize\centering\arraybackslash}X>{\hsize=0.895\hsize\centering\arraybackslash}X@{}}
\toprule
\textbf{Contrast} & \textbf{R\&M, registered (n~=~45)} & \textbf{NRC-VAD, natural (n~=~161)} & \textbf{NRC-VAD, matched (n~=~45)} & \textbf{Warriner, natural (n~=~126)} & \textbf{Warriner, matched (n~=~42)} \\
\midrule
Tuning main effect, valence $\Delta$r (p) & +.056 (.320) & +.058 (.036) & +.056 (.252) & +.062 (.081) & +.052 (.294) \\
Tuning main effect, arousal $\Delta$r (p) & $-$.083 (.357) & $-$.042 (.340) & $-$.071 (.455) & $-$.052 (.393) & $-$.077 (.466) \\
Tuning by scale interaction, $\Delta\Delta$r (p) & +.047 (.944) & +.072 (.548) & +.059 (.874) & +.052 (.762) & +.047 (.918) \\
Arousal tracking with scale, base chain $\Delta$r & +.168 & +.059 & +.145 & +.077 & +.160 \\
Arousal tracking with scale, tuned chain $\Delta$r & +.215 & +.131 & +.204 & +.129 & +.207 \\
Arousal depth-gap growth, base chain $\Delta$z & +.203 & +.104 & +.197 & +.109 & +.167 \\
Arousal depth-gap growth, tuned chain $\Delta$z & +.229 & +.202 & +.251 & +.160 & +.205 \\
Embedding-table valence r, range & +.479 to +.507 & +.504 to +.517 & +.437 to +.479 & +.489 to +.504 & +.508 to +.537 \\
Embedding-table arousal r, range & +.125 to +.179 & +.014 to +.091 & +.201 to +.217 & +.001 to +.039 & +.159 to +.202 \\
\bottomrule
\end{tabularx}
\end{table}

Tuning effects are pooled instruction-tuned minus base at the analysis layer, with nominal p in parentheses. The interaction is the 27B tuning delta minus the 2B tuning delta in raw correlation units, with p from the Fisher z contrast at standard error $\sqrt{4/(n-3)}$. Scale contrasts are 27B minus 2B within a chain, in correlation units for arousal tracking and in Fisher z for depth-gap growth. Embedding-table rows are ranges across the six cells. Table~\ref*{tab:6} gives the verdicts. The reruns under the four additional settings are unregistered.

\begin{table}[H]
\small
\caption{\textbf{The emotion-set control, best component for arousal when the decomposition is refit on the rated subset alone.}}
\label{tab:E.3}
\begin{tabularx}{\linewidth}{@{}>{\hsize=1.690\hsize\raggedright\arraybackslash\hyphenpenalty=10000\exhyphenpenalty=10000}X>{\hsize=0.583\hsize\centering\arraybackslash}X>{\hsize=0.927\hsize\raggedright\arraybackslash}X>{\hsize=0.837\hsize\raggedright\arraybackslash}X>{\hsize=0.927\hsize\raggedright\arraybackslash}X>{\hsize=0.927\hsize\raggedright\arraybackslash}X>{\hsize=0.927\hsize\raggedright\arraybackslash}X>{\hsize=0.927\hsize\raggedright\arraybackslash}X>{\hsize=1.256\hsize\centering\arraybackslash}X@{}}
\toprule
\textbf{Norm source} & \textbf{n} & \textbf{2B base} & \textbf{2B tuned} & \textbf{9B base} & \textbf{9B tuned} & \textbf{27B base} & \textbf{27B tuned} & \textbf{Cells surviving} \\
\midrule
Russell and Mehrabian, registered & 45 & third & third & third & third & third & third & 0 of 6 \\
NRC-VAD, natural & 161 & second & third & second & second & second & second & 5 of 6 \\
NRC-VAD, matched & 45 & third & third & third & third & third & third & 0 of 6 \\
Warriner, natural & 126 & second & third & second & fourth & fourth & fourth & 2 of 6 \\
Warriner, matched & 42 & third & third & third & third & second & second & 2 of 6 \\
\bottomrule
\end{tabularx}
\end{table}

The counts are point estimates. The Russell and Mehrabian row is the registered control, and the other four settings are unregistered. In an unregistered check that re-grades nothing, the rated emotions were resampled 1,000 times per setting, with each draw shared across the six cells. Component assignments vary under resampling in every cell and every setting, the registered one included, and no cell has arousal on the second component in 95 percent or more of resamples, or in 5 percent or fewer. Per cell, arousal lands on the second component in .25 to .36 of resamples on the 45 Russell and Mehrabian emotions and in .23 to .76 across all settings. Each setting's point count recurs in .59 of resamples under Russell and Mehrabian, .07 under NRC-VAD natural, .60 under NRC-VAD matched, .07 under Warriner natural, and .06 under Warriner matched (n = 42). P(0) and P(6) are the proportions of resamples in which arousal's best-correlating component is PC2 in none, or in all six, of the cells. They are .59 and .22 for Russell and Mehrabian, .22 and .42 for NRC-VAD natural, .60 and .18 for NRC-VAD matched, .47 and .25 for Warriner natural, and .36 and .40 for Warriner matched.

\begin{table}[H]
\small
\caption{\textbf{The scale battery.}}
\label{tab:E.4}
\begin{tabularx}{\linewidth}{@{}>{\hsize=1.216\hsize\raggedright\arraybackslash\hyphenpenalty=10000\exhyphenpenalty=10000}X>{\hsize=0.638\hsize\centering\arraybackslash}X>{\hsize=0.638\hsize\centering\arraybackslash}X>{\hsize=0.481\hsize\centering\arraybackslash}X>{\hsize=1.249\hsize\centering\arraybackslash}X>{\hsize=1.778\hsize\raggedright\arraybackslash}X@{}}
\toprule
\textbf{Registered measure} & \textbf{Base chain $\Delta$} & \textbf{Tuned chain $\Delta$} & \textbf{Floor} & \textbf{p, base and tuned} & \textbf{Verdict} \\
\midrule
Valence r (PC1 with valence ratings) & +.0096 & +.0082 & .15 & .928, .927 & null, sub-floor in both chains \\ \addlinespace[3pt]
Arousal r (PC2 with arousal ratings) & +.1683 & +.2153 & .15 & .237, .200 & null after Holm in both chains, above floor \\ \addlinespace[3pt]
PC1 variance share (pp) & $-$4.8848 & $-$4.6192 & 1.0 & \textless{} $10^{-15}$, \textless{} $10^{-15}$ & real in both chains, opposite to the registered direction \\ \addlinespace[3pt]
PC2 variance share (pp) & +0.6286 & +0.1894 & 1.0 & $4.7 \times 10^{-13}$, .029 & null by floor in both chains, with the base-chain rise passing Holm and the tuned-chain rise not \\ \addlinespace[3pt]
Mid-late layer stability & $-$.04240 & $-$.00027 & .01 & n/a & real in the base chain, carried by layer 45 (\S8.3), and null, sub-floor in the tuned chain \\
\bottomrule
\end{tabularx}
\end{table}

$\Delta$ is 27B minus 2B within a chain, in correlation units for the two correlation rows and in percentage points for the two variance rows. Each chain carries the same five registered measures, so the family has ten rows. The eight rows that carry a p form the Holm family (m = 8), and the two mid-late layer-stability rows are graded on floor alone by the case-bootstrap test fixed at grading (\S3, \S8.3) and are printed at the precision of their artifact. The registered scale prediction's second clause, that the gap between PC2's and PC3's shares would widen, was never operationalized as a statistic and is ungraded, and PC3's share was not graded on the scale axis. The two PC1 variance-share p values are printed as a bound, p \textless{} $10^{-15}$, because the computed tail underflows double precision and a normal approximation cannot support a finer claim.

One unregistered extension of \citet[\S6.4]{hollowell2026replicating} to six cells bears on the arousal rise of \S5.3. The rise is confined to the second component. Across all six cells the multiple correlation of arousal on the second and third components together is nearly flat, .725 to .750, while its correlation with the second component alone runs .363 to .668. In the base chain the amount the third component adds falls with size, from +.225 at 2B to +.149 at 9B to +.078 at 27B. This extension re-grades nothing.

\begin{table}[H]
\small
\caption{\textbf{Dependence-aware bootstrap of the tuning correlation contrasts.}}
\label{tab:E.5}
\begin{tabularx}{\linewidth}{@{}>{\hsize=2.194\hsize\raggedright\arraybackslash\hyphenpenalty=10000\exhyphenpenalty=10000}X>{\hsize=0.257\hsize\centering\arraybackslash}X>{\hsize=0.548\hsize\centering\arraybackslash}X@{}}
\toprule
\textbf{Contrast} & \textbf{Pooled $\Delta$} & \textbf{95 percent interval} \\
\midrule
Valence r (PC1 with valence ratings) & +.0558 & +.0303 to +.0910 \\
Arousal r (PC2 with arousal ratings) & $-$.0829 & $-$.1290 to $-$.0446 \\
Off-diagonal, PC1 with arousal ratings & +.0409 & +.0139 to +.0690 \\
Off-diagonal, PC2 with valence ratings & $-$.0881 & $-$.1390 to $-$.0351 \\
Interaction, arousal, 27B minus 2B & +.0470 & $-$.0051 to +.0988 \\
\bottomrule
\end{tabularx}
\end{table}

Each of 2,000 draws samples the 45 rated emotions with replacement and applies the same sample to every cell, so the intervals allow for the emotions the compared correlations share. Each cell's principal components stay as fit on all 171 emotions, and in every draw each axis is re-signed to agree with its human rating, as the registered statistic is on the full set. Pooled $\Delta$ is the mean of the three per-size deltas, and the interval is the 95 percent percentile interval. The analysis grades no contrast, is unregistered, and the per-size values are in the released file.

\textbf{Three tuning measures outside the Holm correction (\S7.1).} The two norm correlations were registered as exploratory, and cohesion was registered with a direction and neither a floor nor a test, so none of the three raises the bar for the primary tests. The two norm correlations relate each emotion's vector norm with its position in the emotion space, over all 171 emotions. Tuning changes the correlation between the norm and the position on the valence axis (PC1) by +.095 pooled, against a .10 floor (p = .043), just under it. It changes the correlation between the norm and the distance from the center of the arousal axis (PC2), the absolute value of that position, by +.045. Cohesion, which has no registered floor, moves +.017. The .10 floor rests on the independence assumption described in \S3, which inflates it by an amount we did not measure.

\textbf{The readability instrument (\S6.2).} For an emotion's cleaned vector $v$ at a given layer, each token's score is its entry in $W_U^{\top}\big((1 + w) \odot v \,/\, \sqrt{\mathrm{mean}(v^2) + \varepsilon}\big)$, where $W_U$ is the unembedding matrix, $w$ the final normalization weight, and $\varepsilon = 10^{-6}$, and the readability test ranks the emotion's own word among those scores.

\textbf{Readability checks (\S6.2).} Three checks ask whether the gate at 27B and the early readability at 2B and 9B belong to the test rather than to the vectors. The redraws of the seam and the 27B onset were pre-stated (\S11 lists the filings). The 2B and 9B layer-0 result under redraws, the removal of the normalization and of the cleanup, and the random-direction control are unregistered. First, the vectors are cleaned against a basis drawn from the neutral dialogues (\S3), and a different draw might move the onset or dissolve it. Across 500 redraws of the cleanup basis, the onset stays at layer 23 in both 27B cells, and the four smaller cells read from layer 0 in every draw. Five of the 500 draws failed to converge under the default solver on the cloud machines that ran them and were computed under the alternative solver. On a draw that converges under both, the two solvers agree to $9.1 \times 10^{-12}$, with no effect on the seam or the onset. On another machine, draw 158 failed to converge again and the fallback gave the same seam, layer 23. The other four converged under the default solver there.

Second, we remove each convention of the test in turn and remeasure. Layer-0 readability at 2B and 9B survives without the normalization in all four cells, and survives without the cleanup in three of the four. The exception is 2B tuned, whose layer-0 median rises to 14,800 on uncleaned vectors, so its early readability is specific to the cleaned vectors. The 27B layer-0 medians stay at chance or worse under the same checks. Third, the three sizes differ in width, 2,304 to 4,608 coordinates, so low ranks at the smaller sizes could in principle be a property of narrower spaces rather than of the emotion vectors. Random directions passed through the same readout score near rank 128,000, chance, at all three widths.

\textbf{The seam threshold (\S6.3).} The consolidation seam is defined at .97 agreement with the analysis layer, held for three layers. Across thresholds from .95 to .98 the smaller cells' seams stay between a seventh and about a third of depth and both 27B seams stay at half, sitting at layer 23 under .95, .96 and .97 and moving one layer, to 24, at .98.

\textbf{The calibration analysis (\S6.3).} We pre-stated one calibration analysis to ask whether the meeting of geometry consolidation and readability onset at layer 23 in the 27B cells is sensitive to where the readability line is drawn. It makes two tests. The onset line of \S6.2 sits at rank 10,000, roughly the top 4 percent of the vocabulary. The first test replaces 10,000 with a rank taken from the cell's own median-rank trajectory across depth (Figure~\ref*{fig:4}), halfway on the logarithmic rank scale between the trajectory's early plateau, averaged over its first five layers, and its late floor, averaged over its last four. The second test tightens the line, replacing 10,000 (top 4 percent) with 1,000 (top 0.4 percent) and 100 (top 0.04 percent).

In the first test the trajectory puts the line at about 2,300 in 27B base and 1,700 in 27B tuned, four to six times stricter than 10,000. The crossing comes at layer 25.47 in 27B tuned and 25.67 in 27B base, a fifth of a layer apart and both about two and a half layers past the seam. The fractional crossing is a linear interpolation, in layer and $\log_{10}$ median rank, between the two sampled layers that bracket the first descent through the line, computed on the reported draw of the cleanup basis without resampling. A stricter line cannot be crossed earlier than a looser one, so the lag's direction follows from the line. We had fixed the comparison in advance. A crossing rounding to layers 22 through 25, one layer either side of the seam's resample interval of 23 to 24, counts as meeting the seam, and 26 or beyond counts as later. The rounding boundary at 25.5 falls between the two cells, so tuned counts as meeting and base as later, and both verdicts sit close to it, tuned by three hundredths of a layer and base by seventeen hundredths. We had fixed both averages in advance. An unregistered check that averages the last five layers instead moves tuned to 25.28 and base to 25.55, changing neither verdict.

In the second test, within the pre-stated window of layers 18 to 28, neither 27B cell reaches 1,000 or 100. Past that window, descriptively, both cells first cross 1,000 at layer 29 and 100 at layer 34, six and eleven layers after the seam. On the reported draw, the one draw of the cleanup basis behind the results we report, the tuned cell sits at rank 1,013 at layer 27, just above the line, and first crosses it at layer 29. Across 500 redraws of the cleanup basis, the tuned cell's distance between seam and onset under the 1,000 line is zero in none of the draws, against 75.6 percent of draws under the 10,000 line. In the redraws where the tuned cell reaches 1,000 inside the window, it crosses at layer 26 or 27 (median 26), two to four layers after the seam. The coincidence survives neither tighter line.

\FloatBarrier
\section{A valence asymmetry in readability}

\citet[\S10.1]{hollowell2026replicating} left open whether the vocabulary readout differs by valence. That study observed that its direct-unembedding readout was cleanest for negative high-arousal emotions and messiest for positive ones, and noted that valence, arousal, and word frequency were entangled in the rows behind the observation. \citet{venkatesh2026negative} asked a related question by a different method: whether language models process positive and negative valence at the same depth. Patching activations in three small tuned models, with short scenarios carrying good or bad news as stimuli, he found they do not. Negative valence localizes to early layers and positive valence peaks at mid-to-late layers, and he left open whether base models show the same asymmetry. The readout of \S6.2 lets us ask a related question of every cell in the family, base and tuned: do positive and negative emotion words become readable at the same depth? This analysis is unregistered and exploratory. The frequency control below was pre-stated.

To split the emotions by valence we use the NRC-VAD Lexicon v2.1 \citep{mohammad2025nrcvad}, a set of human ratings of English words on valence, arousal, and dominance. It rates 137 of the 140 single-token emotion words from \S6.2, 50 positive and 87 negative, and lacks energized, insulted, and stimulated. We then compare the median rank of the positive words with the median rank of the negative words at every layer, in every cell.

The two groups differ at both ends of the network. At layer 0 in every 2B and 9B cell, the positive words are already readable, with medians of 1,000 to 1,900, and the negative words sit well behind them, with medians of 6,700 to 11,300. In the 2B cells the negative median starts above the 10,000 line and crosses it at layer 2 in base and layer 4 in tuned. In the 9B cells it starts below the line, at 7,776 and 6,695, so both groups are readable from layer 0 there and only the difference in rank separates them. At 27B, where neither group is readable until half depth, the positive median crosses below 10,000 at layer 22 in base and layer 21 in tuned, one to two layers before the negative median does at layer 23. By the final layer the order has reversed. The negative words end the more readable in all six cells, with medians of 13 to 25 against 18 to 42 for the positive words.

Word frequency could produce the early asymmetry on its own, since a more common word might be promoted from the start whatever its valence. Three controls preserve the early positive advantage in the four smaller cells. First, the positive and negative word sets do not differ detectably in frequency. Second, with frequency controlled, the positive advantage at layer 0 keeps its direction in all four 2B and 9B cells. Third, subsets of positive and negative words matched on frequency preserve that advantage in the same four cells. In the 27B cells neither group is readable at layer 0, and under the controls the difference between them runs the other way, significantly in the tuned cell (p = .026) and not in the base cell (p = .079). The controls also ran at the final layer. There the controlled comparison puts negative words ahead in every cell, significantly in 9B base, 9B tuned, and 27B base. The positive advantage therefore belongs to the early layers and the negative advantage to the late ones.

Valence shapes readability at every size and under both tunings. Positive emotion words read first in four of the six cells, from layer 0 in the 2B cells and a layer or two ahead of negative words at the 27B gate. The ordering at the gate is observed without a frequency control. In the 9B cells both groups read from layer 0, and the positive words lead on rank alone. Negative emotion words read best by the end. In the 27B cells, the onset of \S6.2 summarizes two valence groups whose medians cross the line one to two layers apart. \S5.1 found valence in the embedding table of every cell, and in the four cells that never gate that finding has a readout counterpart from the first layer. In those four cells the early asymmetry does not depend on word frequency. It is present in base models, which is the half of Venkatesh's question a readout can answer. Which direction his causal asymmetry would take on this family is a question for patching, not for a lens.

\FloatBarrier
\section{The fixed-probe method block}

Companion to \S7.2 through \S7.4. The regime differences from \citet{sofroniew2026emotion} are disclosed here rather than in the body, and the study's one out-of-sample prediction is recorded at the end.

The measurement holds emotion probes fixed and reads how much of each probe's direction the activations carry, before and after tuning. Each probe is an emotion's raw difference-of-means vector from \S3, taken before the cleanup step. For each size we use the tuned cell's probes on both cells, matching the post-trained-probe choice in \citet{sofroniew2026emotion}. Activations come from the 500 shared neutral dialogues, averaged over the tokens from the fiftieth onward in each dialogue and computed at every layer for both cells of each size. The reported quantity is the mean cosine, over the 500 dialogues, between an emotion's probe and the dialogue's pooled activation. \S7.2 uses each size's analysis layer.

\citet{sofroniew2026emotion} differ from this study in token position, prompt mix, and probe extraction. They read one token per prompt, the colon token after ``Assistant,'' on a mix of neutral and challenging prompts, while we average over every token from the fiftieth onward in neutral dialogue text. They apply one probe set, from the post-trained model, to both Claude checkpoints. We do the same at each size, using the tuned cell's probes on both cells, though each cell also has its own extracted probes. The uncentered base and tuned profiles agree at .9967 to .9993, so a probe set fit on one cell reads the other nearly the same, which supports the choice to read both cells with one probe set within these pairs. Sofroniew et al.~report .83 between Claude's base and post-trained profiles on neutral prompts (their Figure 36), against the uncentered .9967 to .9993 of \S7.2, and the two values are not directly comparable for these reasons. The activation results of \S7.2 are therefore an analog of their comparison, not a replication.

\textbf{The Claude comparison (\S7.2).} \citet{sofroniew2026emotion} publish per-emotion shifts for Claude Sonnet 4.5 (their Table 16), averaged over all their prompts at the Assistant colon token. The correlation between our shift profile and theirs is $-$.24 at 2B, +.23 at 9B, and +.03 at 27B, or +.25 at 27B once the shared offset is removed. At 2B, tuning raises spiteful by .023, the fourth largest rise at that size, and Claude lowers spiteful more than any other emotion. Claude's three largest rises are brooding, gloomy, and reflective, and at 27B reflective is among the ten emotions tuning lowers most. Claude leaves regretful, remorseful, guilty, and self-critical near zero (each within .004) and raises sorry by .011, where 27B lowers all five.

A centered version of the activation profiles subtracts the mean activation across the 500 dialogues, per dimension, before the cosines are taken. It gives base-to-tuned correlations of .833 at 2B, $-$.121 at 9B, and .954 at 27B (\S7.2). At 9B the disagreement sits in a few dimensions. We ranked each cell's dimensions by their variance across the dialogues and excluded the union of the two cells' twenty highest, twenty-one in all, which raises the centered correlation to .96. Excluding only the single highest-variance dimension raises it to .77, and the two highest to .92. No random selection of twenty-one dimensions does the same in 200 draws (numpy default\_rng, seed 0). The twenty-one were chosen as the principled set, not the best one: a sweep over the number of excluded dimensions peaks near ten, and one dimension among the twenty-one moves the correlation the other way. The profiles that remain are reliable. When we split the dialogues into odd and even halves and recenter within each half, the two halves' profiles correlate at .975 in the base cell and .983 in the tuned cell. Why a few dimensions can dominate follows from the measurement itself: each cell's profile is the probe set's projection onto a single vector, the mean of the unit-normalized dialogue activations, and centering shrinks that vector's norm from about .94 to about .02, so a handful of high-variance dimensions set the remainder. The dimensions responsible are specific to 9B. This diagnostic does not explain why the 27B shifts diverge from the smaller sizes (\S7.2).

\textbf{The 27B late third (\S7.3).} Removing layer 45 alone drops 27B's late-third perturbation to .0103, below the middle third's .0142. Removing layer 44 alone leaves it at .0165, still above. Removing both leaves .0043, below the early third's .0100.

\textbf{Nulls for the split search (\S7.3).} A search for the best split guarantees some difference between the two sides, so we asked whether the observed difference exceeds what the search alone would produce. We shuffled the layer labels and repeated the search 2,000 times. At 9B and 27B, no shuffle produced a difference as large as the observed one (largest shuffled values .322 and .416). At 2B, one did, at .683 against the observed .620. The observed difference there exceeds all but one of the 2,000 shuffled values (add-one p = .0010, computed as in \S4.3). The null quantifies what the search contributes and grades nothing.

A second null replaces the shift vectors with uncorrelated random ones and reruns the same split search, in 2,000 draws per size. No draw reached an observed difference, the largest null value was .051, and each observed difference ran at least twelve times its own size's null maximum, so the search alone cannot manufacture structure of this magnitude. The claim rests on the pattern across sizes, and we do not interpret where any split falls, since under pure noise the preferred split lands anywhere in the allowed range. The observed difference between same-side and opposite-side mean correlations is .620 at 2B, .386 at 9B, and .605 at 27B. The split falls at layer 16 of 46 at 27B, at layer 27 of 42 at 9B, and at layer 23 of 26 at 2B, so the 2B late group is the last three layers rather than a half of the network.

\textbf{Displacement checks and the regression (\S7.4).} The 1,000 redraws of the cleanup basis behind \S7.4 are a run separate from the 500 redraws behind the readability analyses of \S6 (\S3). Under them, each of the three vindictive emotions is in the top ten of the displacement ranking in at least .999 of redraws at 27B. At 2B and 9B, vindictive and vengeful reach the top ten in .42 to .66 of redraws, and spiteful in only .174 and .070.

Purity is the share of an emotion's ten nearest neighbors, by cosine, that belong to its own family, and its average over the 171 emotions is the cohesion score of Table~\ref*{tab:1}. Per-emotion purity is largely shared across cells, with a mean correlation of .62 between cells, so emotions near family boundaries tend to be similar across cells. The three guilt-related emotions score .85 (regretful), .88 (remorseful), and .87 (guilty), and the three vindictive emotions .60 to .70.

The regression's indicator covers eight self-conscious moral emotions fixed in advance: ashamed, guilty, embarrassed, humiliated, proud, remorseful, regretful, and self-critical. Seven enter the regression because self-critical has no NRC-VAD rating, and the regression runs over the 137 emotions with NRC-VAD ratings and single-token names. In the 27B activation shifts, vindictive, vengeful, and spiteful rank 21st, 25th, and 44th in how far tuning lowers them. Valence magnitude and arousal come from the NRC-VAD lexicon v2.1 \citep{mohammad2025nrcvad}. Layer-0 readability is the log of the rank at which an emotion's own word appears when its layer-0 vector is projected onto the vocabulary (\S6.2), so a higher value means less readable, averaged over the 2B and 9B cells, since emotion words are not readable at layer 0 at 27B. The indicator is a 0/1 term and the other predictors are standardized, so their coefficients are not comparable in size. The indicator's standard error is 1.03 against arousal's 0.09. The model explains 11 percent of the variance in displacement.

\textbf{The out-of-sample prediction.} One prediction in this study was pre-stated separately from the preregistration and tested on data that did not exist when it was filed. Its design was fixed by a dated filing and a dated amendment, not by a hash, so its record is a date rather than a checksum. It drew its direction from the emotions \S7.2 found tuning lowers most at 27B and predicted that tuning lowers the readings of five probes named in advance, at a measurement point locked in advance, on twelve scenarios from the replication study's protocol rather than the neutral dialogues above. Three probes are primary, the three guilt-related emotions of \S7.4 (regretful, remorseful, guilty), and two are secondary (self-critical, reflective). All five moved in the predicted direction, and the one-sided p for the three primary probes together reached .056. Of the primary probes, only regretful individually has a one-sided p below .05, and reflective, a secondary probe, has one at p = .004. The direction transfers from \S7.2 as a matter of sign and not of rank: of the ten emotions \S7.2 finds tuning lowers most and the ten this measurement lowers most, only jealous and nostalgic appear in both, and the three guilt-related emotions rank 35, 56, and 58 of 171 here. The filing named the noise bound, the standard error across scenarios, but left the numeric threshold open. We closed that gap at grading, with the .056 already in view, by adopting a one-sided threshold of p \textless{} .05, which grades our own prediction null.

\FloatBarrier
\section{The layer-45 replication}

Companion to \S8.3. It records the two departures from the registration's wording in the layer-stability measure, the rule under which base 27B's late-layer divergence was replicated, the layer sweep, two further measures that point to the same late band, and the bearing of the depth curves on \citet{zhang2025decoding}.

\textbf{Departures from the registration.} The registration's floor table labels this measure as agreement with the analysis layer, but the floor measurement the registration cites computes the pairwise form reported in \S8.3, and the floor was set on that quantity. The registration also names cosine where its code computes Pearson. Recomputed under cosine, no value moves by more than $4 \times 10^{-5}$ and no verdict changes. A second registered measure of depth stability, retention of each emotion's nearest neighbors across layers, is computed and released but was not graded (Appendix C, C2-11).

\textbf{The replication rule.} The first measurement of the divergence rested on a single full-depth extraction. We therefore re-extracted base 27B at layers 31, 42, and 45, twice with different seeds, after verifying the checkpoint's weight hashes. The decision rule was fixed by a dated filing rather than a hash. A control extraction of layer 31, where base 27B's stored vectors already exist, agreed with them at .9849, the value the rule had predicted, so the pipeline was faithful. The ceiling on that agreement is .9856, the mean over 30 subsample draws of each draw's agreement with the full 1,200-story structure. The rule's first run returned no verdict, because its threshold had assumed that layer 45 agrees with itself as closely as layer 31 does. Measured, it does not. Two extractions of layer 45 agree at only .895, and the square-root rule, which assumes that the two extractions' deviations from the layer's true structure are independent and equal, puts the layer's ceiling at the square root of their agreement, .946. Layer 31's ceiling comes from subsample draws and not from that rule, and the square-root rule gives .9854 there, in agreement with it. The lower ceiling at layer 45 is itself evidence that the layer is more sensitive to the story sample. Under the corrected rule, fixed before the second seed was read, the divergence graded genuine on both seeds.

\textbf{The layer sweep.} A sweep of every layer from 42 to 45 filled in the shape. All four replicate, and the divergence is a slope rather than a cliff, steepest at the single step from layer 43 to 44 (consecutive-layer correlations .9970 into 43, .9087 into 44, .9800 into 45). It is also distributed. The dimension that changes most is 4140, a dimension whose activations run far larger than the rest (Appendix J). It accounts for 13.8 percent of the change, but excluding it, or the top ten or hundred, makes agreement worse rather than restoring it. The tuned 27B cell's late divergence is markedly milder.

\textbf{Two further measures.} Two unregistered measures point to the same late band. Residual norms grow across depth in every cell, and the largest single-step drop in norm falls at the same late transition in both 27B cells and nowhere else. This residual-norm curve is a per-dialogue-mean proxy. It averages norms per dialogue over this study's neutral-dialogue activations, where \citet{hollowell2026replicating} measured them per token, because the per-token norm files are not available. Rotation of individual emotion vectors between adjacent layers concentrates in base 27B's final layers, is absent at 2B and 9B, and in the tuned 27B cell moves to the consolidation seam of \S6.3.

\textbf{The depth curves and \citet{zhang2025decoding}.} \citet{zhang2025decoding} report that emotion representation declines at the final layers of the models they probe. Their quantity is not the divergence. Its counterpart here is the valence correlation at each layer. In this family the decline in that correlation belongs to base 27B, where the valence correlation falls by .24 between layers 43 and 45, most of its total decline of .256, while the tuned 27B cell's valence never declines. Tuning appears to remove base 27B's late-layer fall in the valence correlation, which any account of the mechanism will have to fit.

\FloatBarrier
\section{Licenses for existing assets}

All third-party assets are used in compliance with their terms. This study inherits the replication study's asset set unchanged except for the models, where six checkpoints replace one. The release decisions those terms motivate are stated in \S11.

\begin{table}[H]
\small
\caption{\textbf{Models.}}
\label{tab:I.1}
\begin{tabularx}{\linewidth}{@{}>{\hsize=0.591\hsize\raggedright\arraybackslash\hyphenpenalty=10000\exhyphenpenalty=10000}X>{\hsize=1.667\hsize\raggedright\arraybackslash}X>{\hsize=0.614\hsize\raggedright\arraybackslash}X>{\hsize=1.128\hsize\raggedright\arraybackslash}X@{}}
\toprule
\textbf{Asset} & \textbf{Version} & \textbf{License} & \textbf{Condition relevant to this release} \\
\midrule
Gemma 2, six checkpoints & \texttt{google/gemma-2-2b} (c5ebcd40), \texttt{-2b-it} (299a8560), \texttt{-9b} (33c19302), \texttt{-9b-it} (11c9b309), \texttt{-27b} (938270f5), \texttt{-27b-it} (aaf20e6b), each pinned to the revision shown & Gemma Terms of Use & Every released vector set, analysis-layer and embedding-table, is a Model Derivative and is distributed under the Gemma Terms with the notice and use restrictions. Only raw weight tensors, the vocabulary-readout ingredients, ship as a regeneration recipe rather than as weights \\ \addlinespace[3pt]
Claude Sonnet 4.5 & \texttt{claude-sonnet-4-5-20250929} & Anthropic Commercial Terms & Governs the generated stimulus corpus inherited from the replication study, which ships as a generation recipe \\
\bottomrule
\end{tabularx}
\end{table}

\begin{table}[H]
\small
\caption{\textbf{Norm sources.}}
\label{tab:I.2}
\begin{tabularx}{\linewidth}{@{}>{\hsize=0.816\hsize\raggedright\arraybackslash\hyphenpenalty=10000\exhyphenpenalty=10000}X>{\hsize=0.568\hsize\raggedright\arraybackslash}X>{\hsize=1.045\hsize\raggedright\arraybackslash}X>{\hsize=1.572\hsize\raggedright\arraybackslash}X@{}}
\toprule
\textbf{Asset} & \textbf{Version} & \textbf{License} & \textbf{Condition} \\
\midrule
\citet{russell1977evidence} norms & 45-emotion derived table & no dataset license, data within the published article & Ships as a small derived table of factual normative ratings, with provenance and citation \\ \addlinespace[3pt]
NRC-VAD Lexicon & v2.1 & NRC non-commercial research terms & Redistribution prohibited. Ships as provenance plus fetch script \\ \addlinespace[3pt]
\citet{warriner2013norms} norms & 13,915-lemma set & CC BY-NC-ND 3.0 & Non-commercial, no derivatives. Ships as provenance plus fetch script \\
\bottomrule
\end{tabularx}
\end{table}

The software environment of each machine is pinned in the release manifest.

\textbf{Reproduced source materials.} The 171-emotion vocabulary and the generation prompts originate in the appendix of \citet{sofroniew2026emotion} and reach this study through the replication. They are not reproduced here, and Appendix A of \citet{hollowell2026replicating} carries them.

\FloatBarrier
\section{Compute}

We load each model with TransformerLens 3.3.0's TransformerBridge and leave its compatibility mode off, which keeps the released weights unprocessed: no layer-norm folding, no centering of the writing weights or the unembedding, and no value-bias folding. Extraction ran in bfloat16 on 80 GB accelerators, with the precision and cross-hardware spot-checks of \S8.4 on a second accelerator generation, and the re-extractions behind the late-layer case study of \S8.3 ran the same way. Total accelerator use was approximately 20 hours, most of it in the extraction pass across the six checkpoints. Analysis ran on a Mac in a CPU environment whose packages are pinned in the release manifest. The transport battery and two exploratory analyses ran on the same Mac in a second environment, under Apple Accelerate with single-threaded BLAS, whose package list the release carries as provenance/transport\_battery\_env\_freeze.txt. The permutation floor and the seam bootstrap ran on cloud CPU machines under Linux, whose package lists the manifest also pins.

Reproduction is cheaper than production, which is the point of releasing the vectors. Every result computed downstream of the extracted vectors, meaning the geometry, the invariance tests, the transport battery, the norm-robustness pass, and the depth analyses, recomputes from the released artifacts on a CPU. The activation profiles of \S7.2 and the depth curve of Appendix H read the neutral-dialogue activations rather than the vectors, six per-cell files of about 1.6 GB in all, from which the per-dialogue-mean proxy of Appendix H is computed, and which the released extraction script regenerates from the pinned checkpoints and the released neutral corpus. Reproducing the vocabulary readout of \S6 requires downloading the six checkpoints under a Gemma access grant and exporting their unembedding ingredients, a model download rather than an extraction job, after which those analyses also run locally. The bag-of-words baseline of \S4.3 likewise needs the Gemma tokenizer, which is gated, and the story corpus, which is not released.

The twin run extracted the same 6,840 per-story vectors, 40 for each of the 171 emotions, by two paths on one accelerator. One stopped the forward pass at the layer being read, and the other completed it. Every corresponding vector came back bit-identical before any cross-hardware comparison was read.

\S8.4 found that per-story vectors extracted in bf16 and in fp32 differ very little. At 2B and 9B, the pair with the lowest cosine spreads that difference across many dimensions, with the largest carrying 7 percent of it at 2B and 3 percent at 9B. One dimension at 27B, 4140, carries 97.6 percent. The quantization literature documents this pattern, in which outlier features emerge with scale \citep{dettmers2022llmint8} as massive activations at fixed dimensions \citep{sun2024massive}. \citet{jeong2026shared} reports a model-dependent precision effect on extracted emotion vectors under INT8 quantization, a coarser change than the one measured here.

\FloatBarrier

\end{document}